\documentclass[journal]{IEEEtran}
\usepackage{amsmath,amsfonts}
\usepackage{algorithmic}
\usepackage{algorithm}
\usepackage{array}
\usepackage{textcomp}
\usepackage{stfloats}
\usepackage{url}
\usepackage{verbatim}
\usepackage{graphicx}
\usepackage{cite}
\usepackage{float}
\usepackage{booktabs} 
\usepackage{longtable}
\usepackage{xspace}
\usepackage{tikz}
\usepackage{flushend}
\usepackage{listings}
\usepackage{multicol}
\usepackage{multirow}
\usepackage{adjustbox}
\usepackage{hyperref}
\hypersetup{
	colorlinks=true,
	linkcolor=blue,
	urlcolor=blue,
	citecolor=blue
}
\usepackage{colortbl}
\definecolor{bg}{HTML}{e0f1ff}

\usepackage{makecell}

\definecolor{wood}{RGB}{0,139,0}
\definecolor{veget}{RGB}{0,255,0}
\definecolor{bare}{RGB}{139,0,0}
\definecolor{indus}{RGB}{255,0,0}
\definecolor{resid}{RGB}{205,173,0}
\definecolor{road}{RGB}{83,134,139}
\definecolor{paddy}{RGB}{0,139,139}
\definecolor{planting}{RGB}{139,105,20}
\definecolor{human}{RGB}{189,183,107}

\usepackage{subfigure}
\begin{document}

\title{
Frequency and Edge-Guided Segment Anything Model for Remote Sensing Image Semantic Segmentation
}

\author{
Feng Gao, \emph{Member, IEEE},
Zizhe Pan, 
Haoting Wang, 
Ruzhuang Hua,
Jingchao Cao, \\
Junyu Dong, \emph{Member, IEEE},
Qian Du, \emph{Fellow, IEEE}

\thanks{This work was supported in part by the Natural Science Foundation of Shandong Province under Grant ZR2024MF020, and in part by the Development Program of Shandong Province under Grant 2025CXPT185. (\textit{Corresponding author: Jingchao Cao.})

Feng Gao, Zizhe Pan, Haoting Wang, Ruzhuang Hua, Jingchao Cao, and Junyu Dong are with the State Key Laboratory of Physical Oceanography, Ocean University of China, Qingdao 266100, China.

Qian Du is with the Department of Electrical and Computer Engineering, Mississippi State University, Starkville, MS 39762 USA.}}

\markboth{IEEE Transactions on Geoscience and Remote Sensing}
{Shell}

\maketitle

\begin{abstract}
Remote sensing image semantic segmentation (RSISS) has attracted significant attention due to the growing demand for fine-grained land cover information. The Segment Anything Model (SAM), proposed as a foundation vision model, offers strong segmentation performance and generalization capabilities for RSISS tasks. However, existing SAM-based approaches face two limitations: (1) \textit{Insufficient adaptation of SAM's features to the diverse characteristics of land cover types.} (2) \textit{Semantic ambiguity at object boundaries}, which hinders accurate delineation. To address these limitations, we propose Frequency and Edge-guided SAM (\textbf{FE-SAM}), a scalable and efficient framework for RSISS. Specifically, we introduce a Frequency-Modulated Adapter (\textbf{FMA}) that adaptively decomposes and modulates frequency-domain features based on the input data. It selectively enhances informative high- and low-frequency components corresponding to different land cover types. Furthermore, to improve SAM’s ability to capture fine-grained details, we design \textbf{EGRefiner}, which integrates multi-scale edge-enhanced information extracted from the input image. Extensive experiments on three benchmark datasets demonstrate that FE-SAM outperforms state-of-the-art methods. The source codes are available at:
\url{https://github.com/oucailab/FE-SAM}.

\end{abstract}


\section{INTRODUCTION}

\IEEEPARstart{R}{emote} sensing image analysis is dedicated to processing and interpreting satellite or aerial images, with the goal of offering in-depth understandings of natural environments and human activities. As a pivotal technique, Remote Sensing Image Semantic Segmentation (RSISS) facilitates diverse downstream applications, including natural disaster monitoring \cite{zdp24tgrs}, land cover change detection \cite{zxw24tgrs}, and urban planning \cite{ycy25aaai}. Recent studies have explored the utilization of prior information and model-guided strategies to improve the representation capability of remote sensing images under challenging conditions \cite{Enhanced_DIP_HSI_SR, Model_Informed_Multistage_HSI_SR}.

\begin{figure}[ht]
\centering 
\includegraphics[width=2.5in]{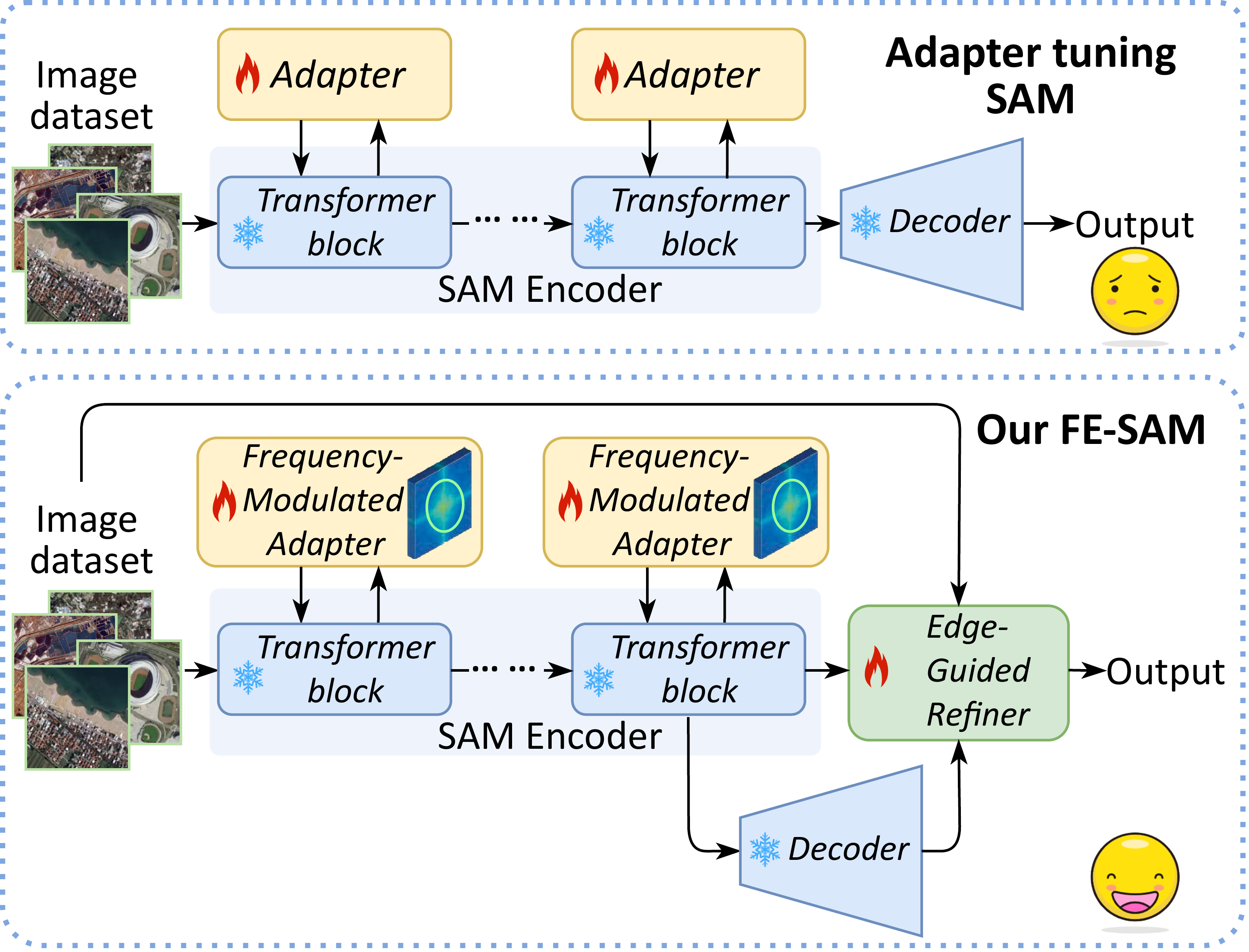}\\ 
\caption{Framework of our Frequency and Edge-guided Segment Anything Model (\textbf{FE-SAM}) for remote sensing image semantic segmentation. It reuses the pre-trained weights of SAM with two novel modules: Frequency-Modulated Adapter (FMA) and Edge-Guided Refiner (EGRefiner).}
\label{fig-mot} 
\end{figure}

In recent years, many deep learning models have been proposed for the RSISS task. Most initial studies leverage the local modeling capacity of convolutional neural networks (CNNs) for segmenting ground objects \cite{hys22grsl, lr21tgrs}. Subsequently, Transformer-based approaches are put forward to capture long-range feature dependencies through the utilization of self-attention mechanisms \cite{mxp24tgrs, lx24grsl}. Although these methods have achieved excellent performance, there is still a gap from practical applications due to the limited network capacity and inadequate high-quality training data.

Recently, the Segment Anything Model (SAM), developed for the purpose of image segmentation, has attracted considerable attention within the computer vision community \cite{sam}. It is trained on a massive dataset of over 11 million images and 1 billion segmentation masks, enabling it to recognize and segment objects across diverse categories. However, SAM is trained on natural images. As there are significant differences in distribution characteristics between the natural and remote sensing images, SAM can hardly be generalized to remote sensing data interpretation. Therefore, in this paper, we aim to explore parameter-efficient fine-tuning of SAM for RSISS task.

It is a non-trivial task for adapting SAM to RSISS task due to the following two challenges: \textbf{(1) \textit{Limitations in adapting SAM's feature to diverse remote sensing land covers.}} SAM is pretrained on general natural images, and remote sensing objects exhibit remarkably different visual characteristics from SAM's training data. In addition, as shown in Fig.  \ref{fig-mot}, different land cover types exhibit distinct spatial distributions, leading to different proportions of high- and low-frequency components. Existing SAM-based remote sensing segmentation methods have limitations in adapting SAM's features to the remote sensing data from a frequency perspective.  \textbf{(2) \textit{Semantic ambiguity of ground object boundaries.}} Natural and man-made backgrounds in remote sensing scenes often contain textures similar to the ground objects of interest. For example, a road may share the same patterns with building's roof, leading to boundary misclassification. In addition, SAM's encoder employs a patch embedding strategy, which inevitably results in the loss of spatial detail information. These two challenges naturally correspond to the two key designs of our model. The first challenge requires adapting SAM features to the scene-dependent spectral characteristics of remote sensing imagery. The second challenge requires recovering fine spatial details degraded in the SAM pipeline.

To tackle these challenges, we propose \textbf{F}requency and \textbf{E}dge-guided \textbf{S}egment \textbf{A}nything \textbf{M}odel (\textbf{FE-SAM}) tailored for RSISS, which integrates \textit{Frequency-Modulated Adapter (FMA)} and \textit{Edge-Guided Refiner (EGRefiner)}. Specifically, FMA is inserted into the frozen SAM encoder to perform energy-guided frequency partitioning and prototype-based spectral recalibration, improving SAM feature adaptation to remote sensing scenes. Meanwhile, EGRefiner incorporates multi-scale edge-enhanced cues from the input image to recover fine spatial details and refine object boundaries. In addition, to enhance fine-grained details of SAM, we propose EGRefiner to incorporate the multi-scale edge-enhanced information from the input image. It can achieve more precise boundary segmentation of ground objects.

Compared with existing SAM adaptation and frequency-aware RSISS methods, \textbf{FE-SAM} focuses on integrating a frequency-aware adapter into the frozen SAM encoder and complementing it with an edge-guided refinement module for boundary recovery. In summary, the main contributions are as follows:

\begin{itemize}

\item \textbf{Model Contribution.} We propose FMA as a lightweight adapter inserted into the frozen SAM encoder. It performs energy-guided frequency partitioning and prototype-based spectral recalibration, enabling parameter-efficient adaptation of SAM features to remote sensing scenes.

\item \textbf{Refinement Contribution.} We design EGRefiner to refine object boundary. It incorporates multi-scale edge-enhanced information from the input image, and can achieve more precise boundary segmentation of desired objects.

\item \textbf{Experimental Contribution.} Extensive experimental validation on three benchmark datasets demonstrates the effectiveness and superiority of our model. We released the codes to facilitate the remote sensing community.

\end{itemize}

\section{RELATED WORKS}

\subsection{Segment Anything Model}

The Segment Anything Model (SAM) \cite{sam}, designed by Meta AI, is a pivotal advancement in the field of foundation models for computer vision, especially segmentation and object detection. It builds on the Vision Transformer \cite{vit} architecture and trains on the 11 million images. SAM has shown remarkable capabilities in zero-shot transfer learning and boasts versatility across many vision tasks, including medical image analysis \cite{yan24wacv, xie24wacv, wei24eccv}, camouflaged object detection \cite{yu24mm, meeran24cvpr, hui24cvpr}, and remote sensing image interpretation \cite{mxp24tgrs, zd24tgrs, shan25cvpr}. To evaluate SAM's generalization ability in complex remote sensing scenes, RSPrompter \cite{rsprompter} uses prompt learning for remote sensing instance segmentation via SAM. RingMo-SAM \cite{ringmosam} constructs an instance-type and terrain-type category-decoupling mask decoder for remote sensing object segmentation. However, these methods fail to enable SAM to adapt to diverse remote sensing land covers, and show poor generalization in complex cases. Different from existing SAM-based methods, FE-SAM studies frequency-domain adaptation under a frozen SAM tuning setting and combines it with edge-guided refinement.

\subsection{Remote Sensing Image Semantic Segmentation}

Deep learning semantic segmentation of remote sensing images has attracted widespread attention. Fully convolutional networks (FCNs) \cite{sww18grsl}, dilated convolution UNet \cite{bc20ijrs}, and attentive bilateral convolution network \cite{abcnet21} have been employed for RSISS. Later, Vision Transformer has been used for remote sensing image analysis, and many hybrid CNN and Transformer networks have been proposed for RSISS \cite{mxp24tgrs, ym24jstars, whl24jstars, xxy24jtstars}. Considering the class-dependent frequency characteristics of remote sensing scenes, as in recent frequency-aware studies\cite{FGSANet, AFENet}, this design enhances SAM’s ability to adapt features to diverse land-cover categories. With the significant progress of large-scale foundation models in computer vision \cite{wjf24cvpr, am25tpami}, foundation models in the field of RS have the potential to solve this problem. RingMo \cite{ringmo22}, SatMAE++ \cite{satmae22, satemaepp24}, and SpectralGPT \cite{Spectralgpt} consider building foundation models for remote sensing images. However, existing SAM-based RSISS methods mainly focus on prompt design, adapter tuning, or decoder modification, while frequency-domain adaptation remains underexplored. Although lightweight Transformer models have studied efficient multi-scale feature interaction\cite{usman2025lightweight}, they are not designed for SAM-based RSISS. Different from these methods and our previous frequency-aware framework, FE-SAM investigates frequency modulation under a frozen SAM adaptation setting and combines it with edge-guided refinement for land-cover representation and boundary delineation. In this paper, leveraging SAM's powerful generalization capability, we can achieve effective semantic segmentation for challenging RSISS task. In addition, we exploit multi-scale edge-enhanced information to refine the object boundary, and the RSISS performance is improved.

\subsection{Edge-Guided Segmentation Refinement}

Accurate boundary delineation is important for dense prediction tasks, especially in remote sensing scenes where adjacent land-cover categories may exhibit similar textures or spectral responses. Edge-guided refinement has been widely used to enhance fine structural details by introducing explicit boundary cues into segmentation models. Recent studies also show that edge-guided or edge-semantics fusion strategies can improve object delineation in complex scenes\cite{EGSAM}.

In SAM-based RSISS, however, fine boundary details may be weakened by patch embedding and lightweight mask decoding. Therefore, besides frequency-domain adaptation, FE-SAM further introduces EGRefiner to extract multi-scale edge-enhanced information from the input image and refine decoder predictions. This design complements FMA by improving spatial detail recovery and boundary quality.

\section{METHODOLOGY}

\begin{figure*} [ht]
\centering 
\includegraphics[width=0.7\textwidth]{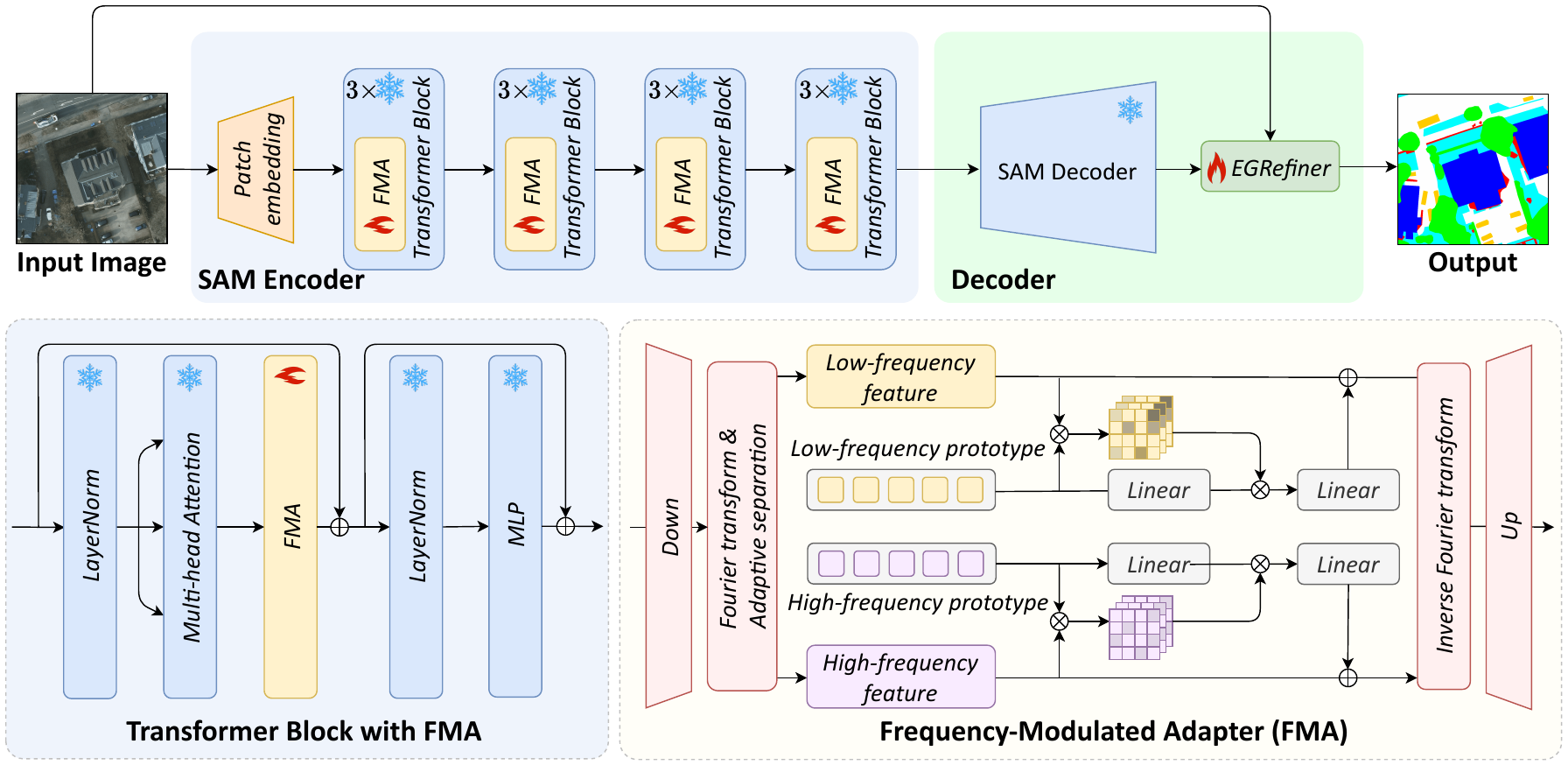}
\caption{Framework of our Frequency and Edge-guided Segment Anything Model (\textbf{FE-SAM}) for remote sensing image semantic segmentation. It reuses the pre-trained weights of SAM with two novel modules: Frequency-Modulated Adapter (FMA) and Edge-Guided Refiner (EGRefiner).}
\label{fig-frame} 
\end{figure*}

In this work, we propose Frequency and Edge-guided Segment Anything Model (FE-SAM)  for the RSISS task. Fig. \ref{fig-frame} shows the overall framework of FE-SAM, which freezes the majority of parameters in the pretrained Vision Transformer encoder, and introduces a small number of parameters through Frequency-Modulated Adapter (FMA). This parameter-efficient design avoids the high computational cost associated with full-model fine-tuning while maintaining the strong representation capability of the pretrained SAM backbone. Recent studies on automated feature engineering have emphasized that effective feature design should balance representation capability and computational efficiency for practical machine learning systems \cite{Auto-Machine}. Inspired by this perspective, FMA performs lightweight frequency-domain feature modulation to selectively adapt SAM features to diverse remote sensing scenes. This design preserves SAM’s original prompt encoding mechanism and enables parameter-efficient adaptation to RSISS. 

In addition, we present the Edge-Guided Refiner (EGRefiner) to enhance the fine-grained details for better segmentation performance. FMA and EGRefiner are integrated in an encoder-decoder framework. FMA adapts encoder features from the frequency perspective, while EGRefiner introduces raw-image edge cues to refine decoder predictions. Their joint optimization enables complementary frequency adaptation and spatial detail recovery. We will detail the Transformer block, FMA, and EGRefiner in the following subsections.

\subsection{Transformer Blocks with FMA}

The SAM encoder contains multiple sequential Transformer blocks. Each Transformer block consists of a multi-head attention layer and an MLP layer. As shown in Fig. \ref{fig-frame}, the FMA is following the multi-head attention and meet a skip connection. In the $i$th Transformer block, the feature extraction in the multi-head attention layer is computed as follows:.
\begin{equation}
    \mathbf{F'}_i=\mathrm{FMA}(\mathrm{Attn}(\mathrm{LN}(\mathbf{F}_i))) + \mathbf{F}_i,
\end{equation}
where $\mathrm{LN}(\cdot)$ denotes the layer normalization, $\mathrm{Attn}(\cdot)$ denotes the self-attention, and $\mathrm{FMA}(\cdot)$ represents the frequency-modulated adapter. Afterwards, the obtained features are fed into the MLP for non-linear feature transformation as follows:
\begin{equation}
    \mathbf{F}_{i+1}=\mathrm{MLP}(\mathrm{LN}(\mathbf{F'}_i))+\mathbf{F}_i,
\end{equation}
here $\mathbf{F}_{i+1}$ is the output features of the $i$th Transformer block and will be fed into the next Transformer block. 

With lightweight FMA, the proposed FE-SAM is capable of aligning SAM's features with the object appearances in remote sensing scenes, and provide reasonable features for the subsequent decoder.

\subsection{Frequency-Modulated Adapter}

\begin{figure} [!ht]
\centering 
\includegraphics[width=3.5in]{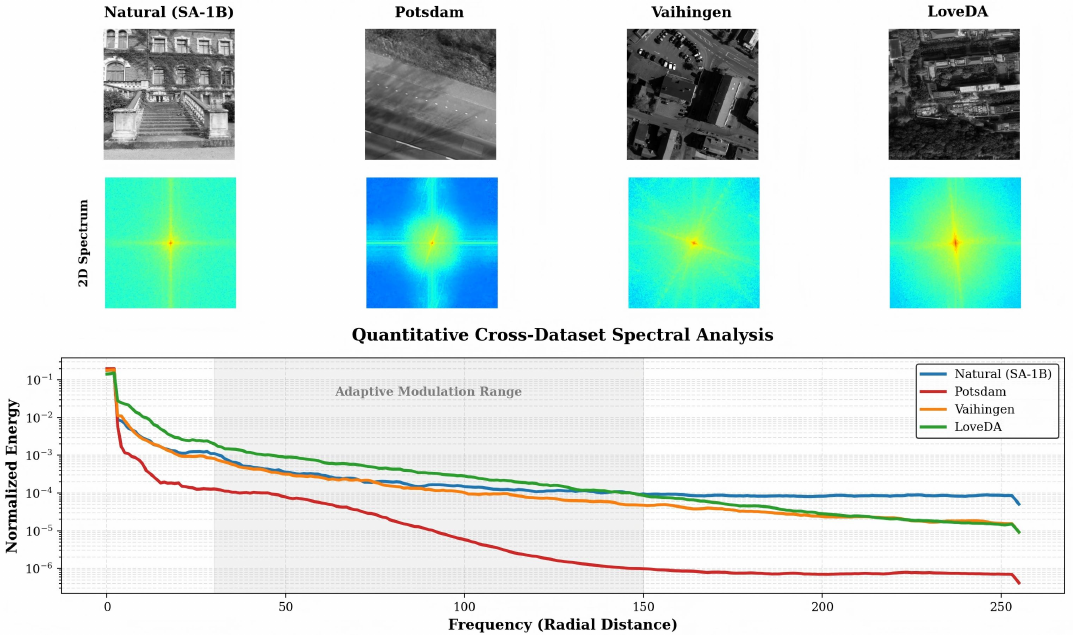}
\caption{Cross-dataset spectral analysis. Top rows: representative patches and their 2D log-magnitude spectra from SA-1B and three remote sensing datasets. Bottom: averaged radial PSD curves.}
\label{QS analysis_result}
\end{figure}

SAM is built upon large-scale natural image datasets. The natural images are commonly with clear foreground-background contrasts and structured object layouts. However, the remote sensing images are generally with complex background and diverse land covers. In addition, different land cover types exhibit distinct spatial distributions, leading to varying proportions of high-frequency and low-frequency components. For instance, urban areas with complex structures (e.g., buildings, roads) generate dense high-frequency signals due to abrupt texture and edge changes. Rural regions predominantly contain low-frequency components from grasslands' smooth and large-scale patterns. Existing SAM-based remote sensing methods mainly adapt SAM through prompts, lightweight adapters, or decoder modifications, but they rarely investigate how the spectral distribution shift between natural and remote sensing images affects SAM feature adaptation.

To solve the problem, we propose Frequency-Modulated Adapter (FMA) to adapt SAM to RSISS. FMA performs input-adaptive frequency partitioning within the frozen SAM encoder and recalibrates the separated spectral components through lightweight prototype-based modulation. This design aims to align SAM features with remote sensing scene statistics under a parameter-efficient tuning setting. As shown in Fig. \ref{fig-frame}, FMA consists of three parts: adaptive frequency separation, frequency feature adaptation, and reconstruction.

To analyze the spectral domain gap, we provide a cross-dataset spectral comparison in Fig. \ref{QS analysis_result}. We compute the averaged radial power spectral density (PSD) using 1,000 randomly selected samples from SA-1B and three RS datasets with different resolutions and regions. As shown in Fig. \ref{QS analysis_result}, natural images exhibit more concentrated spectral responses, while RS images show more dispersed mid-to-high frequency distributions. Despite slight spectral shifts caused by different GSDs and sensors, the three RS datasets share a pattern distinct from the natural domain. This observation suggests a spectral distribution gap between natural and remote sensing images, especially in the mid- and high-frequency ranges. Rather than serving as direct causal evidence, Fig. \ref{QS analysis_result} motivates the frequency-aware design of FMA, which adaptively determines the separation radius $r$ and recalibrates frequency components for RS feature adaptation. Its effectiveness is further verified by the ablation studies in Section IV.D.

\textbf{Adaptive Frequency Separation.} The input data first undergoes a downsampling layer, and then the Fast Fourier Transform (FFT) is employed to convert the feature from spatial domain to the frequency domain. 
Let $\mathcal{M}(u,v)$ denote the magnitude spectrum of the input data, the total energy is the sum of the magnitude-squared spectrum, and can be computed as:
\begin{equation}
\mathbf{E}_t = \sum_{u=0}^{H-1} \sum_{v=0}^{W-1} |\mathcal{M}(u, v)|^2,
\end{equation}
where $H$ and $W$ denote the height and width of the spectrum, respectively. To adaptively modulate the high- and low-frequency components of the input data, FMA determines the frequency separation boundary for each input feature map according to its spectral energy distribution. Although the energy threshold $p$ is fixed as a global hyperparameter, the corresponding radius $r$ is input-dependent. Specifically, we compute the total spectral energy $E_t$ and determine the smallest radius whose accumulated energy from the low-frequency center reaches $p$:

\begin{equation}
r = \min \left\{ \rho \ \bigg| \ 
\frac{\sum_{(u,v)\in \operatorname{Disk}(\rho)} |\mathcal{M}(u, v)|^2}{E_t} \geq p
\right\},
\end{equation}

where $\operatorname{Disk}(\rho)$ denotes a circular region with radius $\rho$ centered at the low-frequency origin of the shifted spectrum. The spectrum inside $\operatorname{Disk}(r)$ is regarded as the low-frequency component, while the remaining part is treated as the high-frequency component. In our experiments, $p$ is set to 0.3, i.e., 30\% of the accumulated spectral energy. Empirical validation shows that this setting provides a better balance between low-frequency structural preservation and high-frequency detail modeling in our experiments.

In the $\mathcal{M}(u,v)$, the spectrum outside the radius $r$ is extracted as high-frequency features, while the spectrum inside the radius $r$ is extracted as low-frequency features. Therefore, we adaptively separate the high- and low-frequency components for remote sensing object feature modeling. The cutoff radius $r$ in FMA is dynamically determined for each image according to the energy criterion, enabling input-adaptive frequency partitioning. This design allows FMA to adjust the separation boundary according to scene-specific spectral distributions, thereby better handling the diverse frequency characteristics of different land-cover types.

In the FMA module, we adopt Fast Fourier Transform (FFT) for frequency decomposition rather than Discrete Wavelet Transform (DWT) or Contourlet Transform. FFT provides an efficient global spectral representation, supports adaptive frequency separation, and enables direct spatial reconstruction through inverse FFT. In contrast, DWT mainly focuses on local multi-scale analysis, while Contourlet Transform introduces higher computational complexity and often requires extra fusion operations. Therefore, FFT is better aligned with the lightweight and globally adaptive design of FE-SAM.

\textbf{Frequency Feature Adaptation.} Now we have low-frequency feature $\mathbf{T}_L$ and high-frequency feature $\mathbf{T}_H$. We introduce two sets of learnable prototypes to distill the key features from the high- and low-frequency components, respectively. For the high-frequency features, we first compute a similarity map $\mathbf{S}_H \in \mathbb{R}^{m\times n}$ between the high-frequency prototype $\mathbf{P}_H\in \mathbb{R}^{m\times c}$ and high-frequency features $\mathbf{T}_H\in \mathbb{R}^{c\times n}$ from the frozen SAM feature, where $m$ denotes the number of frequency prototypes and $c$ is the channel dimension. Here the high-frequency prototype $\mathbf{P}_H$ is a learnable matrix. $\mathbf{S}_H$ is computed as:
\begin{equation}
\mathbf{S}_H= \mathrm{Softmax}(\frac{\mathbf{M}_H \times \mathbf{T}_H}{\sqrt{c}}),
\end{equation}
where $\mathrm{Softmax}$ denotes the softmax activation function, and $c$ represents the channel dimension of the high-frequency features $T_{H}$, which acts as a scaling factor to prevent the dot product from growing too large in magnitude.

In our implementation, $m$ is set to 16 for both branches. Inspired by lightweight model design strategies in practical deployment scenarios\cite{AT_Le_2024}, the prototype number is selected to balance spectral representation capability and computational efficiency. Increasing $m$ may provide finer frequency pattern modeling but introduces additional parameters, while fewer prototypes may limit the representation of diverse land-cover characteristics. Therefore, $m=16$ provides an effective trade-off between adaptation capability and model efficiency.

The prototypes are initialized once before training using offline K-means clustering on randomly sampled low- and high-frequency features after FFT-based adaptive separation. Specifically, the prototype number $m$ is also adopted as the number of clusters in K-means initialization. The resulting $m$ centroids are used to initialize $P_L$ and $P_H$, respectively. During end-to-end training, the prototypes are treated as learnable parameters and updated by standard backpropagation, without dynamic K-means during training or inference. The prototype dimension is kept consistent with the channel dimension of encoder frequency features for stable feature-prototype matching.

Next, we project the prototype $\mathbf{P}_H$ into the feature space of $\mathbf{T}_H$ via a linear layer, followed by the element-wise multiplication with the similarity map $\mathbf{S}_H$. The product with $\mathbf{S}_H$ enables the prototype to better align to $\mathbf{T}_H$, where the high-frequency components from remote sensing data are accentuated. 

Afterwards, we project the obtained features by another linear layer, and fuse them with $\mathbf{T}_H$ via element-wise summation. This process can be denoted as:
\begin{equation}
\hat{\mathbf{T}}_H = W_2((W_1 \mathbf{S}_H) \odot \mathbf{T}_H) + \mathbf{T}_H,
\end{equation}
where $W_1$ and $W_2$ denote the weights of the first and second linear layer, respectively. As shown in Fig. \ref{fig-frame}, the low-frequency features are computed in the same manner as those in the high-frequency branch.

\textbf{Combination and Reconstruction.} The outputs from the high-frequency and low-frequency adaptation are combined. Finally, the combined features are converted to the spatial domain via Inverted Fast Fourier Transform (IFFT), and then undergoes an upsampling layer to generate the output.

\subsection{Edge-Guided Refiner}

While the FMA effectively adapts SAM's features from natural images to remote sensing scenarios, there are still some limitations. First, SAM's encoder employs a patch embedding strategy, which inevitably results in the loss of spatial details. Second, the upsampling approach used in SAM's decoder struggles to recover critical fine-grained information.

\begin{figure}
    \centering
    \includegraphics[width=2.5in]{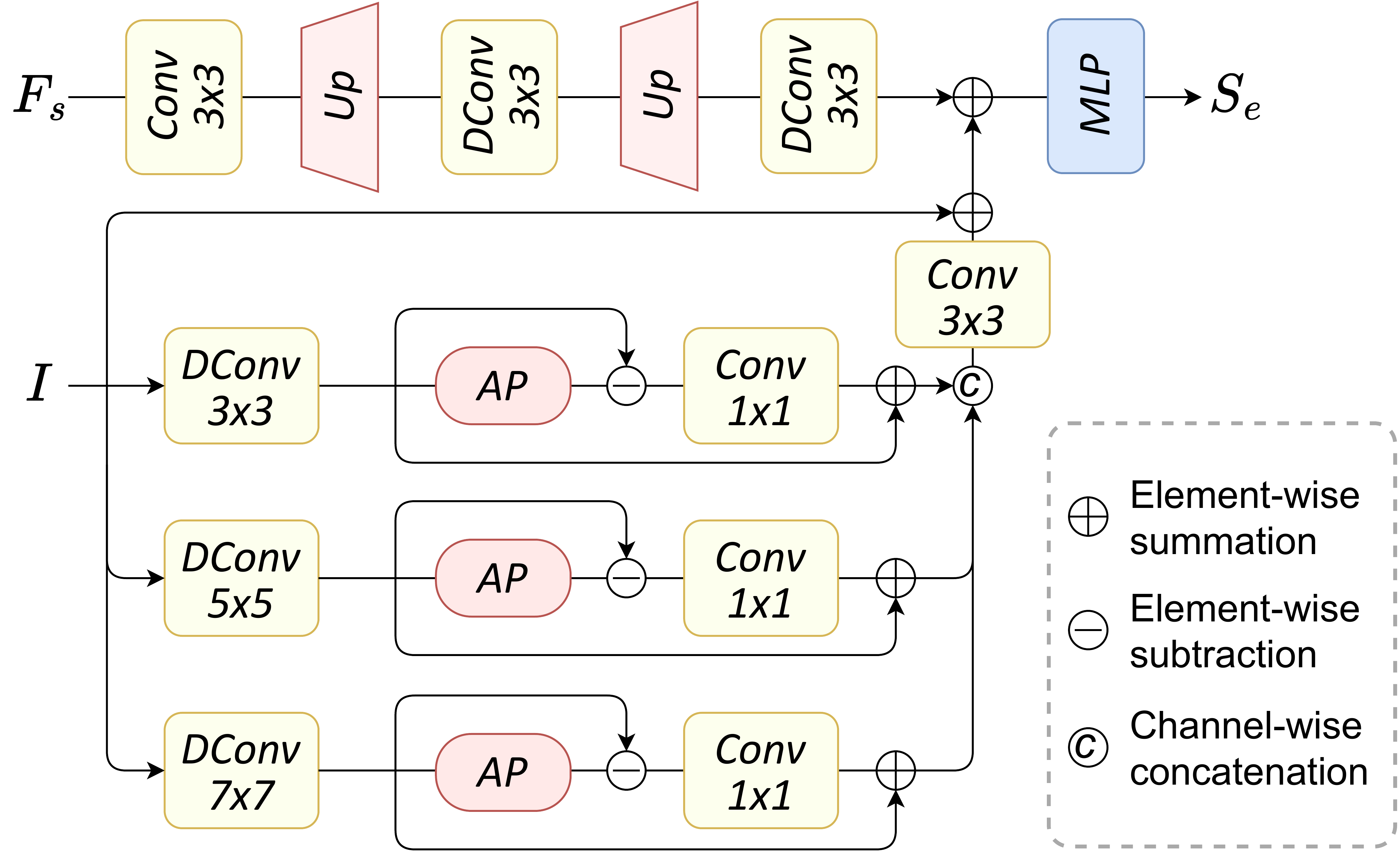}
    \caption{Illustration of the Edge-Guided Refiner (EGRefiner).}
    \label{fig-egr}
\end{figure}

To address these issues, we propose the Edge-Guided Refiner (EGRefiner), as shown in Fig. \ref{fig-egr}. $\mathbf{F}_s$ denotes the feature from the SAM's mask decoder. We use $3\times3$ convolution to extract local features of $F_s$, and then we progressively upsample the features to the input resolution via bilinear interpolations and $3\times3$ convolutions as follows:
\begin{equation}
    \mathbf{F}'_s=\mathcal{C}_{3\times3}(\mathrm{UP}(\mathcal{C}_{3\times3}(\mathrm{UP}(\mathcal{C}_{3\times3}(\mathbf{F}_s))))),
    \label{eq7}
\end{equation}
where $\mathcal{C}_{3\times3}$ is the $3\times3$ convolutional layer, and $\mathrm{UP}$ denotes the $2\times$ upsampling with bilinear interpolation. 

Before feature fusion, $F_s$ is progressively upsampled to the input resolution, where each bilinear interpolation is followed by a $3 \times 3$ convolution for local spatial recalibration. Thus, the decoder feature $F'_s$ and the edge-enhanced feature $F'_{ee}$ have the same spatial resolution before concatenation. This implicit alignment mitigates feature mismatch and enables the fusion of high-level decoder semantics and low-level structural cues.

In order to inject spatial details into $\mathbf{F}'_s$, we extract multi-scale fine-grained details from the input image, rather than from intermediate features produced by the SAM image encoder. This design is motivated by the loss of fine spatial details caused by the patch embedding and downsampling operations in SAM. Compared with intermediate encoder features, the original image retains more complete pixel-level structures for boundary refinement. Moreover, the multi-scale depth-wise convolutions enable EGRefiner to capture boundary details at different spatial scales, which is suitable for remote sensing scenes with objects of diverse sizes and shapes. Specifically, given an input image $I$, we first employ $3\times3$ convolutional layer to extract local features, and obtain $\mathbf{F}_l$. Then, the edge information is extracted through three parallel depth-wise convolutional paths of different scales ($3\times3$, $5\times5$, and $7\times7$) as follows:
\begin{equation}
\mathbf{F}^1_e=\mathcal{C}'_{3\times3}(\mathbf{F}_l), ~~
\mathbf{F}^2_e=\mathcal{C}'_{5\times5}(\mathbf{F}_l), ~~
\mathbf{F}^3_e=\mathcal{C}'_{7\times7}(\mathbf{F}_l),
\end{equation}
where $C'_{3\times3}$, $C'_{5\times5}$, and $C'_{7\times7}$ denote depth-wise convolution with different kernels, respectively. Next, we compute the edge-enhanced features by computing differences between the input and the pooled feature maps as follows:
{
\begin{equation}
\mathbf{F}^i_{ee}=\mathcal{C}_{1\times1}(\mathbf{F}^i_e-\mathrm{AP}(\mathbf{F}^i_e))+\mathbf{F}^i_e, ~~ (1\leq i \leq 3)
\end{equation}
}
where $\mathrm{AP}$ denotes the average pooling with $3\times3$ kernel. Afterwards, we fuse these features with channel-wise concatenation and employ a $3\times3$ convolutional layer as:
\begin{equation}
\mathbf{F}'_{ee}=\mathcal{C}_{3\times3}(\mathrm{Concat}(\mathbf{F}^1_{ee}, \mathbf{F}^2_{ee}, \mathbf{F}^3_{ee})),
\end{equation}
where $\mathrm{Concat}(\cdot)$ denotes channel-wise concatenation. $\mathbf{F}'_{ee}$ contains both fine-grained details and the edge-enhanced information. We use $\mathbf{F}'_{ee}$ to complement the feature $\mathbf{F}'_s$ via element-wise summation. Finally, MLP is employed to generate the final segmentation results $\mathbf{S}_e$. It is computed as follows:
\begin{equation}
\mathbf{S}_e=\mathrm{MLP}(\mathrm{Concat}(\mathbf{F}'_s, \mathbf{F}_l+\mathcal{C}_{3\times3}(\mathbf{F}'_{ee}))).
\label{eq11}
\end{equation}

In Eq. \ref{eq11}, the integrated feature map in EGRefiner has 128 channels and is finally projected by a $1 \times 1$ convolutional layer to produce the probability maps for $N$ target classes.

Multi-scale edge enhancement is performed on the raw input image $I$ rather than on frequency-modulated features. Thus, frequency-adapted semantic features are complemented by fine-grained structural cues from the original image. The final prediction $S_e$ integrates decoder semantic features and EGRefiner-based edge details, enabling complementary frequency adaptation and boundary refinement for more accurate object delineation.

\subsection{Loss Function}

We employ a composite loss function which combines the Binary Cross Entropy (BCE) loss, the Dice loss, and the L1 loss as follows:
\begin{equation}
\mathcal{L}(\mathbf{S}, \mathbf{S}_{gt})=
\mathcal{L}_{BCE}+\mathcal{L}_{Dice}+\mathcal{L}_{L1},
\end{equation}
where $\mathbf{S}_{gt}$ is the ground truth of segmentation. To improve the learning capability, we use $1\times1$ convolutional layer to refine the output of SAM's mask decoder as $\mathbf{S}_m=\mathcal{C}_{1\times1}(\mathbf{F}_s)$. It should be noted that the auxiliary supervision is applied to the dense decoder feature map rather than discrete SAM masks. Specifically, $F_s$ is projected by a $1\times1$ convolution to obtain $S_m \in \mathbb{R}^{H \times W \times N}$, where $N$ denotes the number of semantic classes, including the dataset-specific background or clutter category. Therefore, the auxiliary loss $\mathcal{L}(S_m,S_{gt})$ is computed over the entire image. This dense supervision provides gradients for both foreground and background regions, including areas that may not be well covered by SAM masks, thereby improving the robustness of semantic prediction. Finally, we employ the loss to both the output of SAM's mask decoder and the output of EGRefiner as follows:
\begin{equation}
\mathcal{L}_{total}=\mathcal{L}(\mathbf{S}_e, \mathbf{S}_{gt})+\mathcal{L}(\mathbf{S}_m, \mathbf{S}_{gt}). 
\end{equation}

\section{Experimental Results and Analysis}

To ensure full reproducibility, we implemented FE-SAM using the PyTorch framework. The network was optimized using the AdamW optimizer with an initial learning rate of $1 \times 10^{-4}$ and a weight decay of 0.01. The learning rate was dynamically adjusted using a cosine annealing schedule over a total of 100 epochs. All experiments were conducted on a single NVIDIA RTX 3090 GPU with a batch size of 8.

\subsection{Datasets and Evaluation Metrics}

\textbf{ISPRS Vaihingen}: The Vaihingen dataset consists of 33 very fine spatial resolution TOP image tiles, with an average size of $ 2494 \times 2064 $ pixels. Each TOP image tile has three multispectral bands (near-infrared, red, green), as well as a digital surface model (DSM) and a normalized digital surface model (NDSM), with a ground sampling distance (GSD) of 9 centimeters \cite{dcswin}. The dataset involves five foreground classes (impervious surfaces, buildings, low vegetation, trees, cars) and one background class (clutter). In our experiments, we only used the TOP image tiles and cropped the image tiles into patches of $ 1024 \times 1024 $ pixels.

\textbf{ISPRS Potsdam}: The Potsdam dataset comprises 38 high-resolution TOP image tiles, with an average size of $ 6000 \times 6000 $ pixels \cite{mlc20tgrs}. Each TOP image tile includes four spectral bands (red, green, blue, near-infrared), a digital surface model (DSM), and a normalized digital surface model (NDSM), with a ground sampling distance (GSD) of 5 centimeters. The dataset encompasses six primary classes (impervious surfaces, buildings, low vegetation, trees, cars, clutter). In our analysis, we utilized the TOP image tiles and partitioned them into patches of $ 1024 \times 1024 $ pixels.

\textbf{LoveDA}: The LoveDA dataset\cite{loveda} contains 5987 fine-resolution optical remote sensing images (GSD 0.3 m) at a size of $ 1024 \times 1024 $ pixels and includes 7 landcover categories, i.e. building, road, water, barren, forest, agriculture and background. The dataset encompasses two scenes (urban and rural) which are collected from three cities (Nanjing, Changzhou and Wuhan) in China \cite{song24wacv}. Therefore, considerable challenges are brought due to the multi-scale objects, complex background and inconsistent class distributions.

\begin{table}[htbp]
\centering
\caption{Pixel-level class distribution on three datasets}
\label{tab:class_dis}
  \scalebox{0.85}{
  \begin{tabular}{c|c|cc}
  \hline\toprule
   & Class  & Pixels($\times 10^6$) & Proportion(\%)\\
  \midrule
  \multirow{6}{*}{Vaihingen} &
  Imp. surf.   & 121.5 & 29.2 \\
  & Building     & 113.6 & 27.3 \\
  & Low veg.     & 79.1  & 19.0 \\
  & Tree         & 80.7  & 19.4 \\
  & Car          & 5.0   & 1.2 \\
  & Clutter      & 16.2  & 3.9 \\
  \midrule
  \multirow{6}{*}{Potsdam} & 
  Imp. surf.   & 476.5 & 28.4 \\
  & Building     & 441.3 & 26.3 \\
  & Low veg.     & 322.1 & 19.2 \\
  & Tree         & 265.1 & 15.8 \\
  & Car          & 25.2  & 1.5 \\
  & Clutter      & 147.6 & 8.8 \\
  \midrule
  \multirow{7}{*}{LoveDA} &   
  Background   & 1739.0 & 27.7 \\
  & Building     & 772.2  & 12.3 \\
  & Road         & 634.1  & 10.1 \\
  & Water        & 615.3  & 9.8 \\
  & Barren       & 339.0  & 5.4 \\
  & Forest       & 891.5  & 14.2 \\
  & Agriculture  & 1287.0 & 20.5 \\
  \bottomrule\hline
  \end{tabular}}
\end{table}

The pixel-level class distribution of three datasets are shown in Table \ref{tab:class_dis}. To objectively evaluate semantic segmentation performance, we adopt mean Intersection over Union (mIoU), overall accuracy (OA), and F1 score (F1) as the main evaluation metrics. These metrics provide quantitative comparisons across different methods and datasets, reducing the subjectivity of qualitative visualization. To further assess boundary segmentation quality, we introduce Boundary IoU (B-IoU)\cite{BoundaryIou} as a supplementary metric. Unlike standard mIoU, B-IoU evaluates pixels within a distance $d$ from object boundaries. We set $d=8$ and $d=16$ pixels to measure the delineation accuracy of fine-grained structures, such as building corners and narrow roads.

\subsection{Comparison with State-of-the-Art Methods}

\begin{figure}
    \centering
    \includegraphics[width=0.8\linewidth]{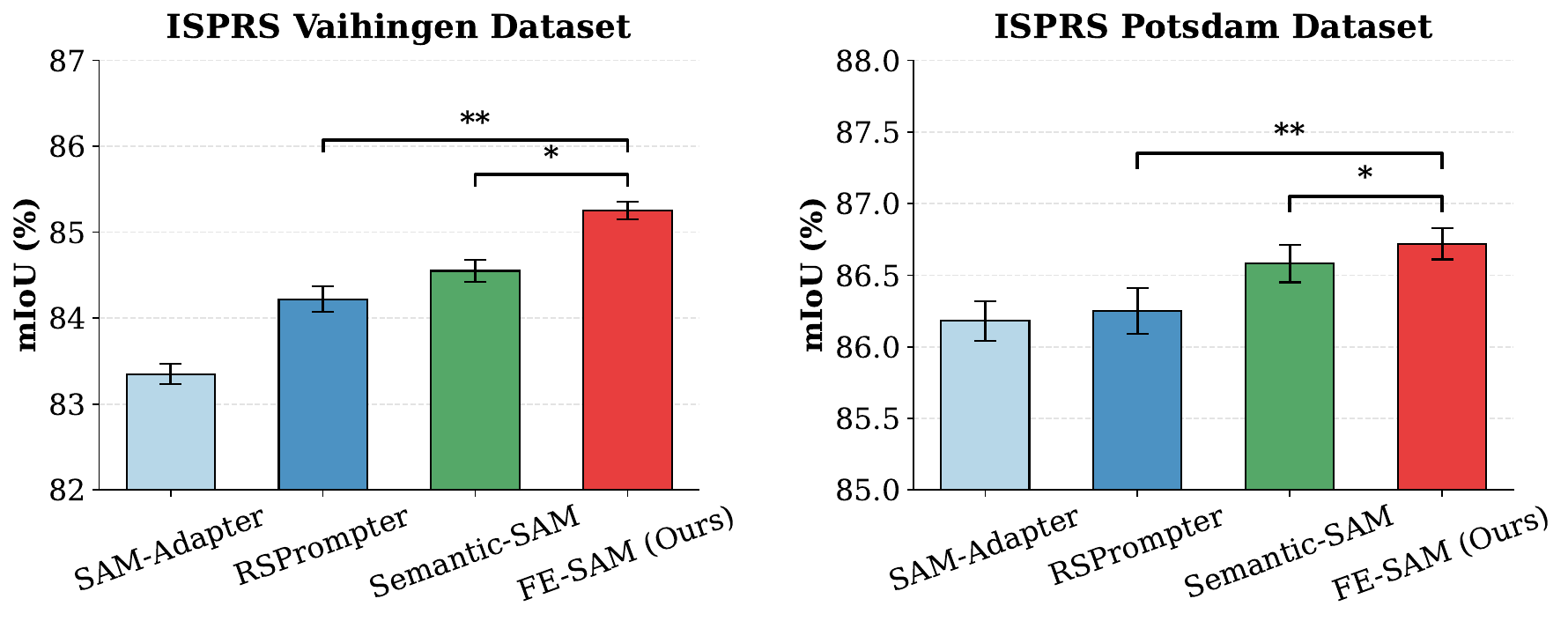}
    \caption{Statistical significance analysis of mIoU on the ISPRS Vaihingen and Potsdam datasets.}
    \label{Statistical significance analysis_result}
\end{figure}

To comprehensively evaluate the effectiveness of our proposed FE-SAM method, we conducted extensive comparisons against a wide array of representative state-of-the-art models. These competitors span three primary categories: 1) CNN-based methods, including FCN8s\cite{fcn}, UNet \cite{unet}, DeepLabv3+ \cite{DeepLabv3+}, MA-Net \cite{MA-Net}, and A2-FPN \cite{A2-FPN}. 2) Transformer-based segmentation methods, including UNetFormer \cite{UNetFormer}, AFENet \cite{AFENet}, CMTFNet \cite{CMTFNet}, FGSANet \cite{FGSANet}, SFFNet \cite{SFFNet}, and SparseFormer \cite{SparseFormer}. 3) Recent SAM-based methods, including the original SAM \cite{sam}, EGSAM \cite{EGSAM}, SAM-Adapter \cite{sam-adapter}, SAMed  \cite{samed}, RSPrompter \cite{rsprompter}, and Semantic-SAM \cite{Semantic-SAM}. 

\begin{table}[htbp]
\centering
\caption{Semantic segmentation results (measured by mIoU) of various methods on three benchmark datasets. The \textbf{bold} and \underline{underline} denote the best and second results.}
\setlength{\tabcolsep}{2pt} 
\scalebox{0.9}{
\begin{tabular}{c|c|ccc}
    \toprule
    \textbf{Method} & \textbf{Type} & \textbf{Vaihingen} & \textbf{Potsdam} & \textbf{LoveDA} \\
    \midrule
    FCN8s & \multirow{5}{*}{\makecell{CNN-\\based}} & 78.25 & 81.03 & 48.02 \\
    U-Net &  & 80.96 & 83.96 & 50.46 \\
    DeepLabv3+ &  & 81.78 & 84.02 & 51.45 \\
    MA-Net &  & 82.76 & 84.91 & 51.64 \\
    A2-FPN &  & 82.40 & 85.16 & 52.20 \\
    \midrule
    UNetFormer & \multirow{6}{*}{\makecell{Transformer-\\based}} & 82.86 & 85.24 & 52.44 \\
    AFENet &  & 84.27 & 86.43 & \underline{54.83} \\
    CMTFNet &  & 83.45 & 85.72 & 52.80 \\
    FGSANet &  & 84.04 & 86.28 & 54.40 \\
    SFFNet &  & 84.10 & 85.78 & 54.42 \\
    SparseFormer &  & 83.83 & 85.87 & 54.17 \\
    \midrule
    SAM & \multirow{7}{*}{\makecell{SAM-\\based}} & 83.04 & 85.94 & 52.54 \\
    EGSAM &  & 83.69 & 86.12 & 54.10 \\
    SAM-Adapter &  & 83.35 & 86.18 & 53.75 \\
    SAMed &  & 83.17 & 86.21 & 54.39 \\
    RSPrompter &  & 84.23 & 86.26 & 54.48 \\
    Semantic-SAM &  & \underline{84.56} & \underline{86.58} & 54.59 \\
    \cellcolor{bg}\textbf{FE-SAM (Ours)} &  & \cellcolor{bg}\textbf{85.24} & \cellcolor{bg}\textbf{86.73} & \cellcolor{bg}\textbf{55.66} \\
    \bottomrule
    \end{tabular}}
\label{tab:comp_all}
\end{table}

\begin{table*}[!ht]
\centering
\caption{Quantitative comparisons with the state-of-the-art methods on the ISPRS Vaihingen dataset. The \textbf{bold} and \underline{underline} denote the best and second results.}
\scalebox{0.9}{
\setlength{\tabcolsep}{4pt} 
\begin{tabular}{c|c|c|ccccc|ccc}
\toprule
\multirow{2}{*}{\textbf{Method}} & \multirow{2}{*}{\textbf{Type}} & \multirow{2}{*}{\textbf{Backbone}} 
& \multicolumn{5}{c|}{\textbf{Per-class F1-score}} 
& \multirow{2}{*}{\textbf{mIoU(\%)}} 
& \multirow{2}{*}{\textbf{m-F1(\%)}} 
& \multirow{2}{*}{\textbf{OA(\%)}}\\
\cmidrule(lr){4-8}
& & & Imp.surf. & Building & Low veg. & Tree & Car & & & \\
\midrule
FCN8s & \multirow{5}{*}{\makecell{CNN-\\based}} & VGG-16 
& 95.22 & 91.89 & 81.76 & 88.52 & 79.68 & 78.25 & 87.41 & 88.96 \\
U-Net & & ResNet-18 & 96.03 & 94.51 & 82.58 & 88.73 & 83.76 & 80.96 & 89.12 & 89.94 \\
DeepLabv3+ & & ResNet-18 & 96.38 & 95.13 & 82.88 & 88.86 & 84.84 & 81.78 & 89.62 & 90.50 \\
MA-Net & & ResNet-18 & 96.39 & 95.17 & 83.86 & 89.51 & 86.33 & 82.76 & 90.25 & 90.69 \\
A2-FPN & & ResNet-18 & 96.31 & 95.04 & 83.01 & 89.08 & 86.67 & 82.40 & 90.02 & 90.52 \\
\midrule
UNetFormer & \multirow{6}{*}{\makecell{Transformer-\\based}} & ResNet-18 & 96.55 & 95.24 & 83.36 & 89.11 & 87.34 & 82.86 & 90.32 & 90.74 \\
AFENet & & ResNet-18 & 96.72 & 95.47 & 84.82 & 90.38 & \underline{89.12} & 84.27 & 91.30 & 91.41 \\
CMTFNet & & ResNet-50 
& 96.49 & 95.43 & 83.96 & 89.48 & 88.04 & 83.45 & 90.68 & 90.88 \\
FGSANet & & FGSABackbone-L & 96.95 & 95.57 & 85.83 & 89.04 & 88.36  & 84.04 & 91.15 & 93.50 \\
SFFNet & & ConvNext-tiny 
& 96.63 & 95.32 & 84.51 & 89.88 & 89.00 & 84.10 & 91.07 & 91.10 \\
SparseFormer & & Swin-tiny 
& 96.46 & 95.39 & 84.16 & 89.70 & 88.58 & 83.83 & 90.86 & 90.96 \\
\midrule
SAM & \multirow{7}{*}{\makecell{SAM-\\based}} & ViT-B 
& 96.83 & 95.24 & 85.17 & 89.53 & 85.84 & 83.04 & 90.52 & 93.14 \\
EGSAM & & ViT-B & 96.78 & 95.15 & 85.47 & 89.77 & 87.53 & 83.69 & 90.94 & 93.40 \\
SAM-Adapter & & SAM-B & 96.90 & 95.00 & 84.67 & \underline{90.41} & 86.61 & 83.35 & 90.72 & 93.25 \\
SAMed & & SAM-B & 96.77 & 95.21 & 84.75 & 90.35 & 85.93 & 83.17 & 90.60 & 93.19 \\
RSPrompter & & SAM-B & \underline{97.06} & 95.72 & \underline{85.93} & 90.19 & 87.44 & 84.23 & 91.26 & 93.56 \\
Semantic-SAM & & SAM-B & 97.04 & \underline{95.81} & 85.66 & 90.22 & 88.58 & \underline{84.56} & \underline{91.46} & \underline{93.57} \\
\cellcolor{bg}\textbf{FE-SAM(Ours)} & & \cellcolor{bg}SAM-B & \cellcolor{bg}\textbf{97.10} & \cellcolor{bg}\textbf{96.03} & \cellcolor{bg}\textbf{86.10} & \cellcolor{bg}\textbf{90.61} & \cellcolor{bg}\textbf{89.55} & \cellcolor{bg}\textbf{85.24} & \cellcolor{bg}\textbf{91.88} & \cellcolor{bg}\textbf{93.79} \\
\bottomrule
\end{tabular}}
\label{tab:comp_vaihingen}
\end{table*}

\begin{figure*} [!ht]
    \centering 
    \includegraphics[width=0.7\textwidth]{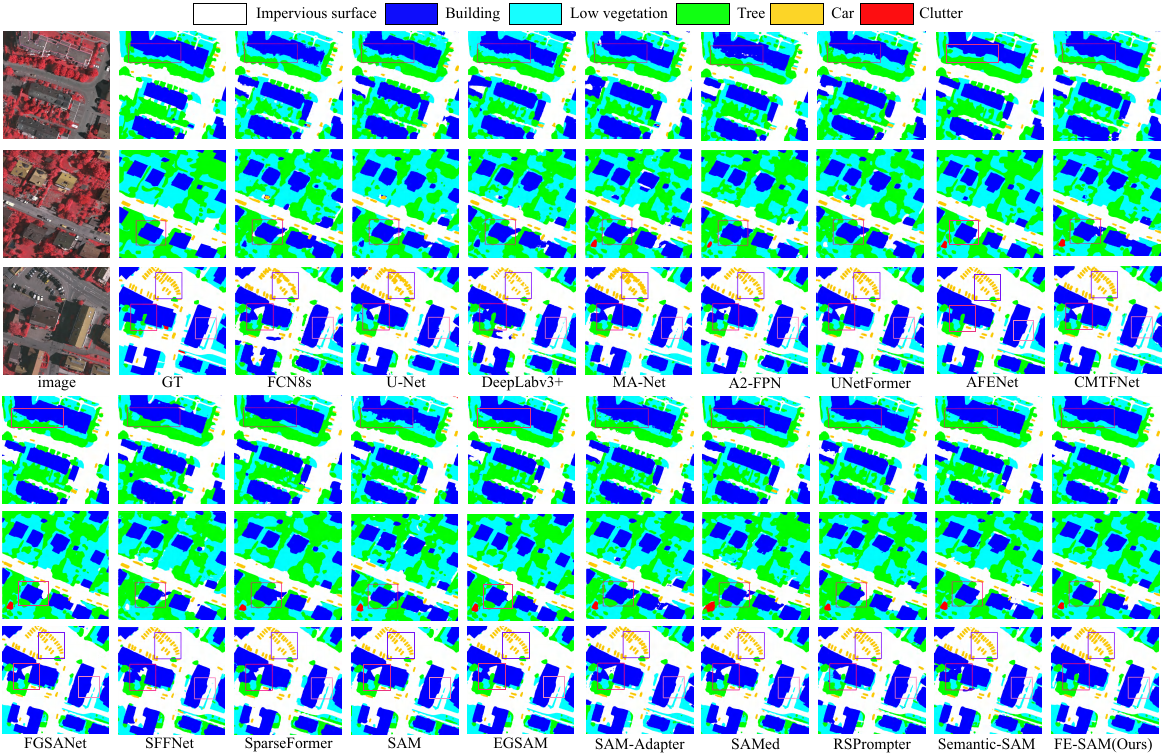} \\ 
    \caption{Additional qualitative results on the ISPRS Vaihingen dataset. From top to bottom, each row showcases a different challenging scene. Our model, FE-SAM (last column), consistently produces segmentation masks that are more detailed and accurate than those from competing methods.}
    \label{vaihingen_result} 
\end{figure*}

\begin{table*}[!ht]
\centering
\caption{Quantitative comparisons with the state-of-the-art methods on the ISPRS Potsdam dataset. The \textbf{bold} and \underline{underline} denote the best and second results.}
\scalebox{0.9}{
\setlength{\tabcolsep}{4pt}
\begin{tabular}{c|c|c|ccccc|ccc}
\toprule
\multirow{2}{*}{\textbf{Method}} & \multirow{2}{*}{\textbf{Type}} & \multirow{2}{*}{\textbf{Backbone}} 
& \multicolumn{5}{c|}{\textbf{Per-class F1-score}} 
& \multirow{2}{*}{\textbf{mIoU(\%)}} 
& \multirow{2}{*}{\textbf{m-F1(\%)}} 
& \multirow{2}{*}{\textbf{OA(\%)}} \\
\cmidrule(lr){4-8}
& & & Imp. surf. & Building & Low veg. & Tree & Car & & & \\
\midrule
FCN8s & \multirow{5}{*}{\makecell{CNN-\\based}} & VGG-16 
& 90.62 & 92.55 & 85.02 & 85.95 & 92.48 & 81.03 & 89.32 & 88.00 \\
U-Net & & ResNet-18 & 92.31 & 94.56 & 85.82 & 87.37 & 95.01 & 83.96 & 91.01 & 89.60 \\
DeepLabv3+ & & ResNet-18 & 92.78 & 95.07 & 86.10 & 87.41 & 94.86 & 84.02 & 91.24 & 89.70 \\
MA-Net & & ResNet-18 & 92.86 & 95.30 & 86.47 & 87.77 & 95.45 & 84.91 & 91.57 & 90.41 \\
A2-FPN & & ResNet-18 & 93.08 & 95.79 & 86.53 & 88.00 & 94.84 & 85.16 & 91.65 & 90.44 \\
\midrule
UNetFormer & \multirow{6}{*}{\makecell{Transformer-\\based}} & ResNet-18 & 92.96 & 95.77 & 86.84 & 87.96 & 95.30 & 85.24 & 91.77 & 90.33 \\
AFENet & & ResNet-18 & 93.74 & 96.22 & 87.75 & 88.63 & \textbf{96.22} & 86.43 & \underline{92.51} & 91.27 \\
CMTFNet & & ResNet-50 & 93.55 & 95.88 & 87.23 & 88.01 & 95.56 & 85.72 & 92.05 & 90.76 \\
FGSANet & & FGSABackbone-L & 93.84 & 96.43 & 87.67 & \underline{88.64} & 95.97 & 86.28 & 92.49 & 91.32 \\
SFFNet & & ConvNext-tiny & 93.48 & 95.91 & 87.40 & 88.15 & 95.61 & 85.78 & 92.11 & 90.88 \\
SparseFormer & & Swin-tiny & 93.43 & 95.95 & 87.30 & 88.24 & 95.90 & 85.87 & 92.16 & 90.84 \\
\midrule
SAM & \multirow{7}{*}{\makecell{SAM-\\based}} & ViT-B & 93.71 & 96.28 & 87.55 & 88.31 & 95.73 & 85.94 & 90.32 & 91.01 \\
EGSAM & & ViT-B & 93.83 & 96.34 & 87.62 
& 88.46 & 95.85 & 86.12 & 92.42 & 91.25 \\
SAM-Adapter & & SAM-B & 93.85 & 96.35 & 87.66 & 88.53 & 95.87 & 86.18 & 92.45 & 91.21 \\
SAMed & & SAM-B & 94.00 & 96.42 & 87.55 & 88.49 & 95.88 & 86.21 & 92.47 & 91.28 \\
RSPrompter & & SAM-B & \underline{94.02} & 96.44 & \underline{87.77} & 88.45 & 95.82 & 86.26 & 92.50 & \underline{91.41} \\
Semantic-SAM & & SAM-B & 94.01 & \underline{96.46} & 87.42 & 88.39 & 96.11 & \underline{86.58} & 92.48 & 91.37 \\
\cellcolor{bg}\textbf{FE-SAM(Ours)} & & \cellcolor{bg}SAM-B & \cellcolor{bg}\textbf{94.03} & \cellcolor{bg}\textbf{96.48} & \cellcolor{bg}\textbf{88.31} & \cellcolor{bg}\textbf{88.93} & \cellcolor{bg}\underline{96.15} & \cellcolor{bg}\textbf{86.73} & \cellcolor{bg}\textbf{92.78} & \cellcolor{bg}\textbf{91.56} \\
\bottomrule
\end{tabular}}
\label{tab:comp_potsdam}
\end{table*}

\begin{figure*} [!ht]
    \centering 
    \includegraphics[width=0.7\textwidth]{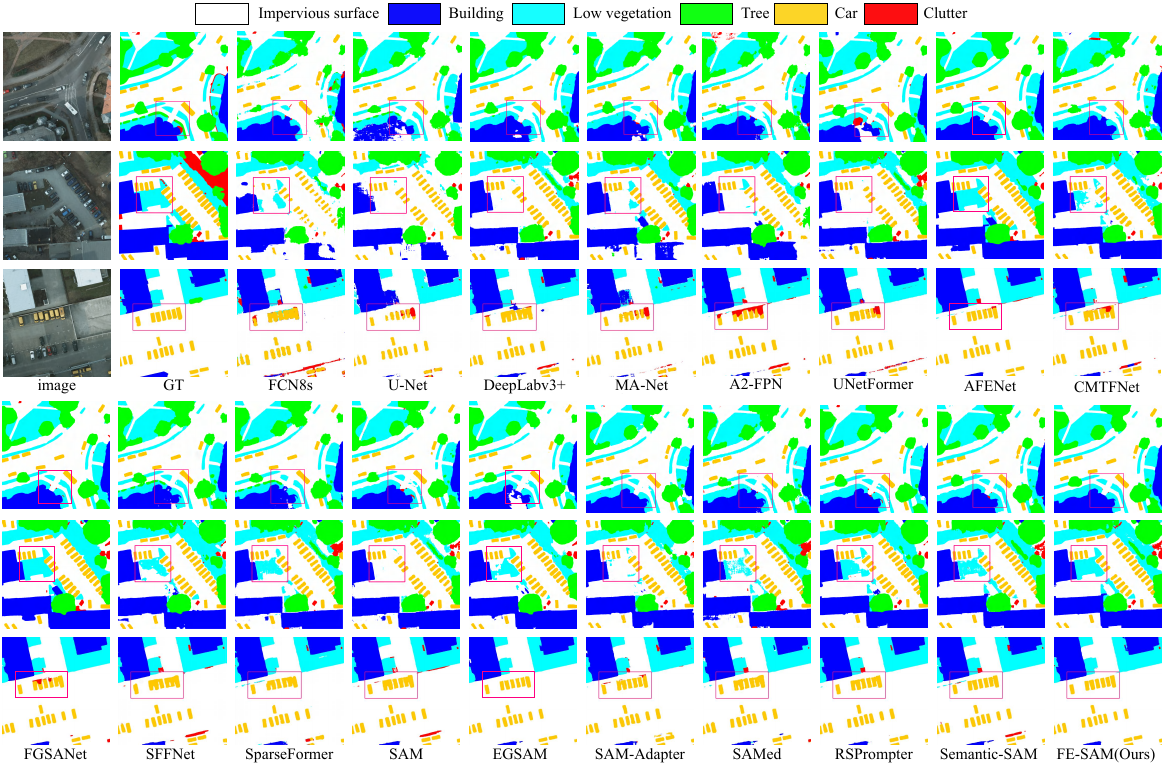} \\ 
    \caption{Additional qualitative results on the ISPRS Potsdam dataset. Our model, FE-SAM (last column), consistently generates more accurate and complete segmentation masks compared to other state-of-the-art methods, especially in areas with complex object layouts.}
    \label{potsdam_result}
\end{figure*}

\begin{table*}[ht]
\centering
\caption{Quantitative comparisons with the state-of-the-art methods on the LoveDA dataset. The \textbf{bold} and \underline{underline} denote the best and second results.}
\scalebox{0.9}{
\setlength{\tabcolsep}{3.5pt} 
    \begin{tabular}{c|c|c|ccccccc|c} 
        \toprule
        \multirow{2}{*}{\textbf{Method}} & \multirow{2}{*}{\textbf{Type}} & \multirow{2}{*}{\textbf{Backbone}} & \multicolumn{7}{c|}{\textbf{Per-class IoU}} & \multirow{2}{*}{\textbf{mIoU(\%)}} \\
        \cmidrule(lr){4-10}
        & & & Background & Building & Road & Water & Barren & Forest & Agriculture & \\
        \midrule
        FCN8s & \multirow{5}{*}{\makecell{CNN-\\based}} & VGG-16 & 54.46 & 63.95 & 56.88 & 62.66 & 11.83 & 32.30 & 47.03 & 48.02 \\
        U-Net & & ResNet-18 & 54.16 & 58.56 & 57.19 & 57.93 & 27.31 & 41.54 & 56.53 & 50.46 \\
        DeepLabv3+ & & ResNet-18 & 54.39 & 60.15 & 58.09 & 63.30 & 26.52 & 41.94 & 55.38 & 51.45 \\
        MA-Net & & ResNet-18 & 54.76 & 60.94 & 57.10 & 62.18 & 34.82 & 37.26 & 54.42 & 51.64 \\
        A2-FPN & & ResNet-18 & 55.64 & 62.00 & 58.42 & 65.56 & 28.08 & 42.19 & 53.54 & 52.20 \\
        \midrule
        UNetFormer & \multirow{6}{*}{\makecell{Transformer-\\based}} & ResNet-18 & 55.73 & 59.68 & 56.81 & 66.36 & 30.36 & 41.36 & 56.75 & 52.44 \\
        AFENet & & ResNet-18 & 47.43 & 59.22 & \textbf{59.17} & \textbf{81.55} & 21.36 & \textbf{48.68} & \textbf{66.40} & \underline{54.83} \\
        CMTFNet & & ResNet-50 & 55.95 & 60.46 & 58.16 & 68.66 & 29.48 & 41.76 & 55.10 & 52.80 \\
        FGSANet & & FGSABackbone-L & 55.90 & 64.20 & 57.80 & 67.80 & 35.50 & 43.80 & 55.80 & 54.40 \\
        SFFNet & & ConvNext-tiny & 56.44 & 65.00 & 58.45 & 66.33 & 32.97 & 44.05 & 57.71 & 54.42 \\
        SparseFormer & & Swin-tiny & \underline{56.59} & 63.03 & 56.73 & 66.32 & 32.30 & 43.85 & \underline{60.36} & 54.17 \\
        \midrule
        SAM & \multirow{7}{*}{\makecell{SAM-\\based}} & ViT-B & 53.66 & 59.41 & 57.91 & 66.91 & 32.50 & 39.74 & 57.66 & 52.54 \\
        EGSAM & & ViT-B & 55.80 & 63.80 & 57.50 & 67.50 & 35.00 & 43.50 & 55.60 & 54.10 \\
        SAM-Adapter & & SAM-B & 55.18 & 63.10 & 57.70 & 67.98 & 34.55 & 43.27 & 54.48 & 53.75 \\
        SAMed & & SAM-B & 56.03 & 63.30 & 58.14 & 67.33 & \underline{36.01} & 43.81 & 56.09 & 54.39 \\
        RSPrompter & & SAM-B & 56.02 & \underline{65.72} & 56.88 & 68.88 & 33.38 & 44.24 & 56.20 & 54.48 \\
        Semantic-SAM & & SAM-B & 55.94 & 65.61 & 58.55 & 68.40 & 33.19 & 43.87 & 56.55 & 54.59 \\
        \cellcolor{bg}\textbf{FE-SAM(Ours)} & & 
        \cellcolor{bg}SAM-B & \cellcolor{bg}\textbf{56.77} & \cellcolor{bg}\textbf{66.10} & \cellcolor{bg}\underline{58.92} & \cellcolor{bg}\underline{69.14} & \cellcolor{bg}\textbf{36.18} & \cellcolor{bg}\underline{44.35} & \cellcolor{bg}58.15 & \cellcolor{bg}\textbf{55.66} \\
        \bottomrule
    \end{tabular}}
    \label{tab:comp_loveda}
\end{table*}

\begin{figure*} [ht]
    \centering 
    \includegraphics[width=0.7\textwidth]{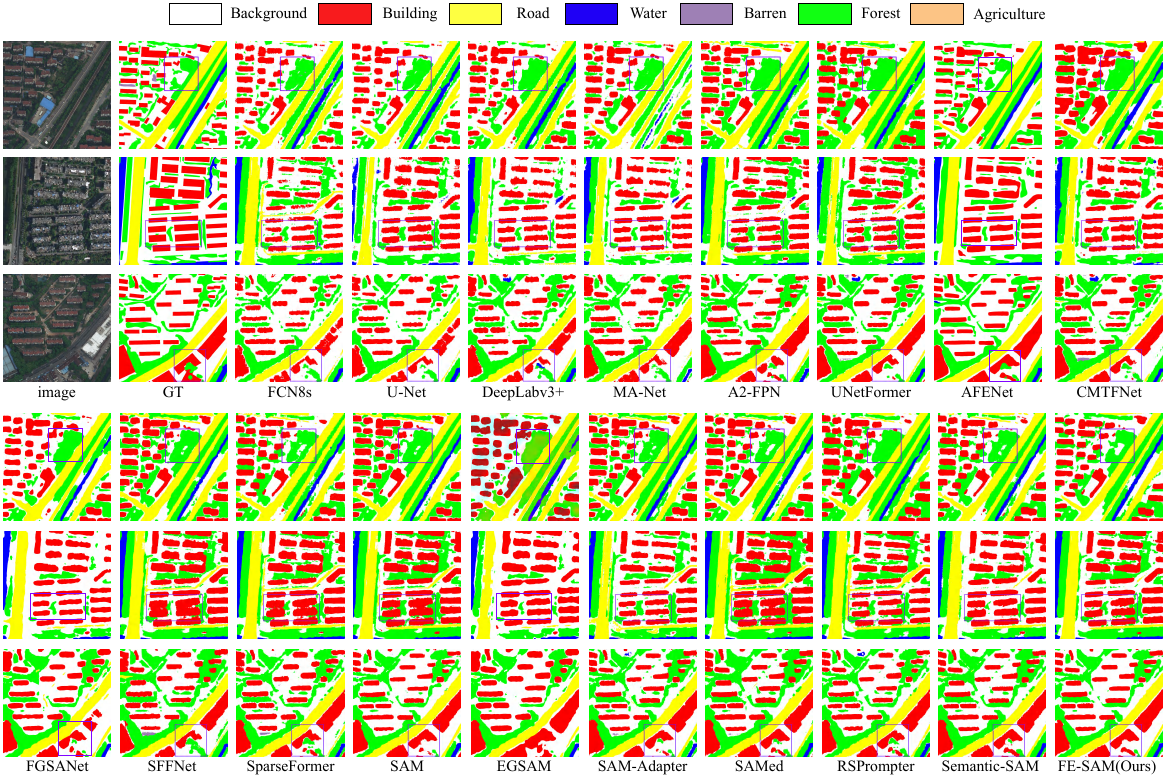} \\  
    \caption{Additional qualitative results on the LoveDA dataset. The selected scenes showcase diverse urban and rural landscapes. Our model, FE-SAM (last column), consistently provides more complete and semantically accurate segmentation masks than competing methods.}
    \label{loveda_result}
\end{figure*}

\textbf{Quantitative Evaluation.} The overall semantic segmentation performance, measured by mIoU, is summarized in Table \ref{tab:comp_all}. The results unequivocally demonstrate that recent SAM-based methods, including our FE-SAM, consistently outperform traditional CNN- and Transformer-based approaches, primarily due to the highly generalized feature representation from SAM's large-scale pretraining. Our FE-SAM surpasses all other methods, achieving the best mIoU scores across all three benchmarks, which validates the overarching efficacy of our proposed frequency-domain adaptation and edge-guided refinement strategy. To provide a more granular, dataset-specific analysis, we present detailed per-class results.

On the ISPRS Vaihingen dataset, our FE-SAM's superiority is most pronounced on man-made objects requiring precise boundary delineation (Table \ref{tab:comp_vaihingen}). For instance, FE-SAM improves the F1-score for `Building' to 96.03\%, surpassing the next-best SAM-based competitor, and significantly outperforms on `Car' with the score of 89.55\%, a notable +0.97\% improvement over the runner-up. This is a direct consequence of our EGRefiner module, which is specifically engineered to counteract the spatial detail loss inherent in SAM's architecture by integrating multi-scale edge information, thereby ensuring high-fidelity geometric reconstruction.

On the ISPRS Potsdam dataset, the per-class F1-scores show that our FE-SAM again achieves top-tier performance  (Table \ref{tab:comp_potsdam}). While excelling on structured classes like `Building', its strength is also evident in resolving ambiguity between spectrally similar classes. Our model achieves the best F1-scores for `Low vegetation' and `Tree', outperforming strong competitors like SparseFormer. This highlights the critical contribution of our FMA module. By empowering the model to differentiate land cover types based on their inherent frequency profiles, the FMA effectively mitigates the semantic confusion that plagues methods relying solely on spectral features.

On the LoveDA dataset, the per-class IoU results in Table~\ref{tab:comp_loveda} confirm the robust generalization of our FE-SAM, which obtains the highest mIoU of 55.66\%. The efficacy of our approach is evident in its balanced performance. It achieves state-of-the-art IoU on large, continuous structures like `Road' and `Water', while simultaneously outperforming most competitors on dense objects like `Building'. This balanced superiority is a direct testament to the FMA module's ability to selectively modulate frequency components—enhancing low-frequencies to preserve the coherence of large-scale structures and boosting high-frequencies to capture the details of small-scale objects.

To assess the stability of FE-SAM, we conducted ten independent trials for each method using random seeds. All trials followed the same data splits, training schedule, augmentation strategy, and evaluation protocol, while only the random seed was changed. For statistical testing, the results obtained under the same random seed were treated as paired observations, and a two-sided paired t-test was conducted on the mIoU differences between FE-SAM and the strongest competing methods. As shown in Fig. \ref{Statistical significance analysis_result}, the error bars denote the standard deviation of mIoU values across trials. In addition, we report the mean mIoU, standard deviation, 95\% confidence interval, and exact $p$-value to provide a more complete statistical comparison. The results show that FE-SAM achieves consistent improvements over the competing methods across different random seeds. The statistical test further indicates that these improvements are unlikely to be caused solely by random seed variations under the current experimental setting. Though our overall metric gains over competing methods are moderate, such improvements carry practical value for remote sensing segmentation, as they boost land-cover mapping and boundary extraction. Consistent performance gains across datasets verify that our frequency adaptation and edge refinement effectively mitigate feature shifts and blurry boundaries.

\textbf{Qualitative Evaluation.} To complement the quantitative metrics, we provide a qualitative analysis to visually demonstrate the segmentation performance of FE-SAM. On the ISPRS Vaihingen dataset (Fig. \ref{vaihingen_result}), it shows the critical role of the EGRefiner module in ensuring high-fidelity geometric reconstruction. As highlighted by the bounded regions in the third row, existing SAM-based methods like SAM-Adapter and SAMed struggle with boundary precision, producing masks with notably blurred edges and rounded corners that fail to adhere to the true object geometry. The sharp, right-angled corners of the building are incorrectly rendered as smooth curves. In contrast, our FE-SAM renders these boundaries with exceptional sharpness and geometric precision, accurately capturing the crisp footprint of the building in a manner that closely mirrors the ground truth. It confirms that by integrating multi-scale edge-enhanced information, the EGRefiner directly counteracts the loss of spatial detail inherent in SAM's architecture.

On the ISPRS Potsdam dataset (Fig. \ref{potsdam_result}), the visualizations expose the limitations of other methods in handling areas with low textural contrast, where severe semantic confusion is a prevalent failure mode. As illustrated in the highlighted region of the second row, a critical failure is the misclassification of large swathes of `Low vegetation' as `Impervious Surface'. Many methods, from classic U-Net to advanced SAM-based adapters, produce an almost entirely uniform and incorrect prediction in this area, failing to capture the presence of the vegetation. Furthermore, the third row highlights the common struggle to preserve discrete `Cars' amidst background noise, where models like U-Net and A2-FPN incorrectly merge the vehicles into the `Clutter' class. Our FE-SAM, however, demonstrates exceptional robustness against both types of errors. It correctly identifies the `Low vegetation' in the second row and delivers a much cleaner segmentation of the individual vehicles in the third row. This comprehensive improvement stems directly from the FMA module. By analyzing and adaptively modulating the distinct frequency profiles, it successfully resolves the spectral ambiguity between low-texture classes like vegetation and impervious surfaces.

On the LoveDA dataset (Fig. \ref{loveda_result}), it highlights FE-SAM's superior generalization. In the second and third rows, competing approaches like A2-FPN and the base SAM produce fragmented road networks where topological connectivity is compromised, and they generate over-smoothed residential areas where individual building footprints are lost or blurred into an indistinct mass. Our FE-SAM, however, exemplifies exceptional adaptability. This is a direct testament to the FMA's capacity to selectively modulate frequency components based on scene content. It can enhance low-frequency components to preserve the coherence of large, continuous structures like roads, while concurrently boosting high-frequency components to capture the intricate details of small, dense building clusters. This scene-specific frequency modulation allows our model to excel at both tasks simultaneously.

\begin{table}[!ht]
\centering
\caption{Quantitative comparison of boundary segmentation quality on the ISPRS Vaihingen dataset. B-IoU ($d=8$) and B-IoU ($d=16$) measure the accuracy within 8 and 16 pixels from the boundary}
\label{table:boundary_iou}
\scalebox{0.9}{
\begin{tabular}{l|c|ccc}
\hline\toprule
\multirow{2}{*}{Method} & \multirow{2}{*}{Backbone}& \multirow{2}{*}{mIoU (\%)} & B-IoU & B-IoU \\
& & & {($d=8$)} & {($d=16$)} \\
\midrule
U-Net[42] & ResNet-18 & 80.96 & 73.12 & 75.48 \\
DeepLabv3+[43] & ResNet-18 & 81.78 & 74.56 & 76.82 \\
UNetFormer[46] & ResNet-18 & 82.86 & 76.28 & 78.53 \\
SAM[8] & ViT-B & 83.04 & 75.82 & 77.96 \\
SAM-Adapter[50] & SAM-B & 83.35 & 77.14 & 79.21 \\
Semantic-SAM[52] & SAM-B & \underline{84.56} & \underline{79.43} & \underline{81.67} \\
\rowcolor{bg}\textbf{FE-SAM (Ours)} & SAM-B & \textbf{85.24} & \textbf{82.15} & \textbf{84.32} \\
\bottomrule\hline
\end{tabular}}
\label{res-Bmiou}
\end{table}

\begin{table}[!ht]
\centering
\caption{Model complexity comparison of SAM-based methods.}
\scalebox{0.9}{
\setlength{\tabcolsep}{3pt}
\begin{tabular}{l|cccc}
\hline\toprule
Method & Total Params(M) & FLOPs(M) & {FPS} \\
\midrule
SAM [8] & 86.94 & 103.17 & {15.6} \\
SAM-Adapter [50] & 90.78 & 109.61 & {14.1} \\
RSPrompter [18] & 93.45 & 113.66 & {11.8} \\
Semantic-SAM [52] & 101.12 & 115.83 & {10.2} \\
\rowcolor{bg}\textbf{FE-SAM (Ours)} & 94.52 & 114.11 & {12.4} \\
\bottomrule\hline
\end{tabular}}
\vspace{2pt}

\footnotesize
\textit{Note}: The parameters of the frozen prompt encoder are included in all complexity statistics for fairness.
\label{tab:complex}
\end{table}

\begin{figure} [!ht]
    \centering 
    \includegraphics[width=2in]{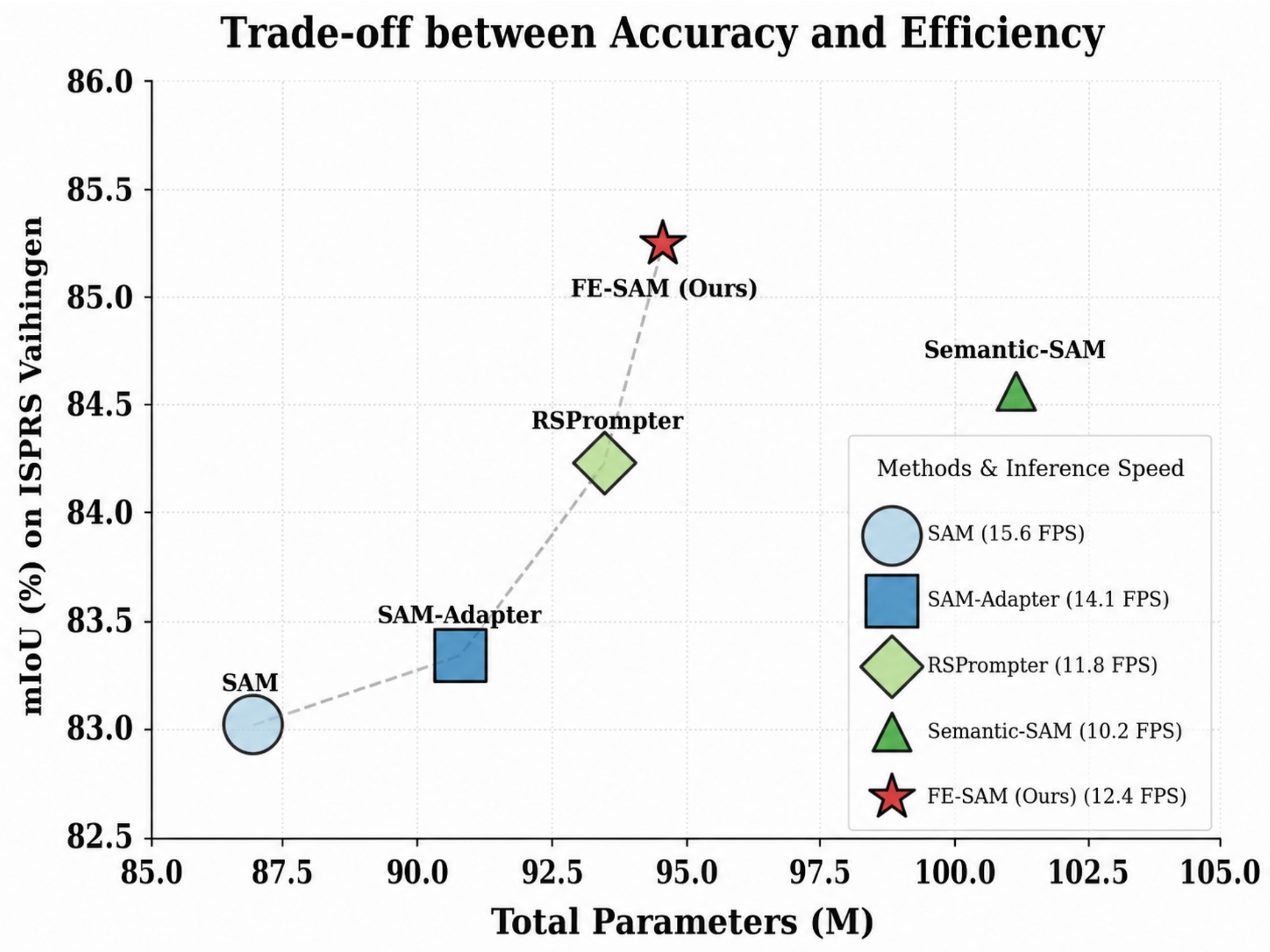}
    \caption{Comparison of mIoU, total parameters, and inference speed among SAM-based methods.}
    \label{efficiency_tradeoff_result}
    \vspace{-1em}
\end{figure}

\textbf{Evaluation of Boundary Precision.} To further evaluate the boundary-sensitive segmentation quality of FE-SAM, we selected several representative models for a comparative study using the Boundary IoU (B-IoU) metric. FE-SAM does not use any boundary-specific loss. Nevertheless, since EGRefiner introduces multi-scale edge-enhanced cues from the original image, we report Boundary IoU (B-IoU) as an auxiliary metric to evaluate boundary-sensitive segmentation quality. B-IoU is used only for evaluation and is not involved in training. As presented in Table \ref{res-Bmiou}, FE-SAM achieves the highest B-IoU scores across different dilation widths ($d = 8$ and $d = 16$) on the Vaihingen dataset. Compared to the state-of-the-art Semantic-SAM, our model provides a more significant margin of improvement in B-IoU than in global mIoU. For instance, the B-IoU ($d = 8$) of FE-SAM reaches 82.15\%, which is 2.72\% higher than the second-best method. These results indicate that EGRefiner improves boundary-sensitive segmentation quality, especially for structures such as building corners and narrow roads.

\textbf{Model Complexity Comparison.} We analyze the complexity of FE-SAM to access its practical viability, and the experimental results are shown in Table \ref{tab:complex} and Fig. \ref{efficiency_tradeoff_result}. Our method introduces a modest overhead of only 7.58 M trainable parameters (mere 8.7\% increase over the original SAM), which yields a substantial performance gain (+2.2 mIoU on ISPRS Vaihingen). While this represents a slightly larger parameter budget than simpler methods like SAM-Adapter, the significant performance improvement validates the efficacy of our specialized FMA and EGRefiner modules, which are tailored to the physical properties of remote sensing data. The FFT and IFFT operations in FMA are computationally efficient on modern hardware, and the EGRefiner follows a bilateral fusion philosophy to balance spatial refinement and inference speed. FE-SAM achieves a competitive inference speed of 12.4 FPS. The efficiency of our FE-SAM is most evident when compared to the strong competitor, Semantic-SAM. FE-SAM not only achieves superior accuracy but does so more efficiently, requiring ~6.6 M fewer parameters and ~1.7 G fewer FLOPs. It shows that our state-of-the-art performance is a result of an intelligent and targeted architecture rather than simply scaling model size. 

\subsection{Ablation Studies}

To validate the contributions of different components in our FE-SAM, we conducted a series of ablation studies. This subsection interrogates the efficacy of our core modules, the architectural principles of the FMA, and the structural integrity of the EGRefiner. All ablations are performed on the challenging ISPRS Vaihingen dataset.

\begin{table}[htbp]
\centering
\caption{Ablation studies on different modules of FE-SAM.}
\scalebox{0.9}{
\begin{tabular}{cc|ccc}
\toprule
\textbf{FMA} & \textbf{EGRefiner} & \textbf{Vaihingen} & \textbf{Potsdam} & \textbf{LoveDA}  \\
\midrule
 & & 83.04 & 85.94 & 52.54 \\
 \checkmark & & 84.06 & 86.50 &54.13 \\
 &\checkmark & 83.57 & 86.25 & 53.64\\
 \rowcolor{bg}\checkmark & \checkmark & \textbf{85.24} & \textbf{86.73} & \textbf{55.66} \\
\bottomrule
\end{tabular}}
\label{tab:ab-1}
\end{table}

As presented in Table \ref{tab:ab-1}, applying the FMA module alone improves the baseline mIoU from 83.04\% to 84.06\%, a significant gain confirming that our frequency-domain adaptation is highly effective at aligning SAM's features with remote sensing data. Similarly, integrating the EGRefiner alone boosts performance to 83.57\%. When both modules are used in our FE-SAM, the performance reaches 85.24\%. It demonstrates that FMA and EGRefiner are not redundant but are highly complementary, working in synergy to address the distinct and orthogonal challenges of feature adaptation (FMA) and boundary refinement (EGRefiner), respectively.

\begin{figure} [!ht]
    \centering 
    \includegraphics[width=2in]{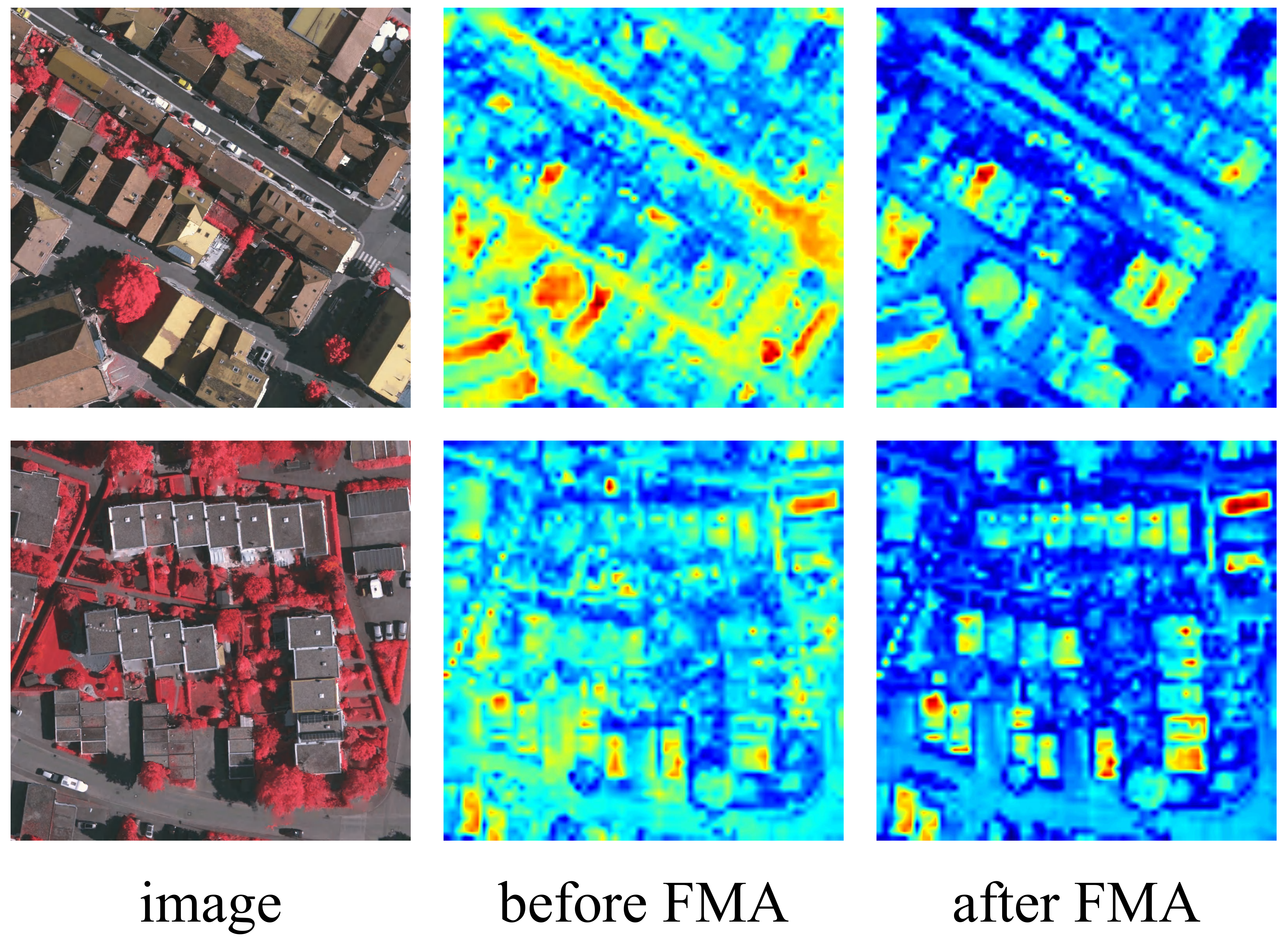}\\ 
    \caption{Visual comparisons of feature representations before and after the FMA module. From left to right: original remote sensing images, feature heatmaps before FMA, and feature heatmaps after FMA.}
    \label{fma_visualization}
\end{figure}

To intuitively illustrate how the FMA module facilitates feature adaptation, Fig. \ref{fma_visualization} visualizes the deep feature maps before and after FMA refinement. For each representative remote sensing patch, we present the original image alongside its corresponding feature heatmaps. The raw feature responses before FMA (middle column) are relatively diffuse, heavily entangled with redundant low-frequency activations from complex backgrounds (e.g., roads and flat terrain). Conversely, after applying FMA (right column), this background noise is significantly suppressed. The refined responses sharply converge onto discriminative object structures and high-frequency geometric boundaries. This visual evidence demonstrates that FMA successfully recalibrates frequency-domain representations, effectively enhancing the adaptability and target-awareness of SAM features in dense remote sensing scenes.

\subsection{Analysis of the Frequency-Modulated Adapter (FMA)} 

\begin{table}[h]
\centering
\caption{Comparison of different parameter-efficient fine-tuning strategies. Performance is measured in mIoU(\%). }
\setlength{\tabcolsep}{3pt} 
\scalebox{0.9}{
\begin{tabular}{c|cccc}
\toprule
\textbf{Method} & \makecell{\textbf{Trainable}\\\textbf{Params (M)}} & \textbf{Vaihingen} & \textbf{Potsdam} & \textbf{LoveDA}\\
\midrule
Full tuning & 83.02 & 83.04 & 85.94 & 52.54 \\
Adapter & 7.83 & 83.12 & 86.09 & 53.14  \\
LoRA & 7.09 & 83.20 & 86.15 & 53.07 \\
\rowcolor{bg} FMA (Ours) & \textbf{7.56} & \textbf{84.06} & \textbf{86.50} & \textbf{54.13} \\
\bottomrule
\end{tabular}}
\label{tab:ab-2}
\end{table}

\textbf{Comparison with Different PEFT Strategies.} To compare the proposed FMA with other parameter-efficient fine-tuning (PEFT) strategies, we conducted a series of experiments, and the results are listed in Table \ref{tab:ab-2}. It can be seen that our FMA consistently achieves the best performance across all three datasets. While LoRA and generic Adapters also offer parameter efficiency, our FMA surpasses them with significantly higher mIoU scores using a comparable or even smaller budget of trainable parameters. It demonstrates that the FMA is an efficient PEFT method for high-resolution remote sensing image segmentation. It achieves favorable trade-off between efficiency and accuracy.

\begin{table}[hbt!]
\centering
\caption{Ablation on the necessity of FMA's adaptive partitioning, evaluated across all three benchmark datasets. Performance is measured in mIoU(\%).}
\scalebox{0.9}{
\begin{tabular}{l|ccc}
\toprule
\textbf{FMA Configuration} & \textbf{Vaihingen} & \textbf{Potsdam} & \textbf{LoveDA} \\
\midrule
Baseline (w/o FMA)       & 83.57 & 86.25 & 53.64 \\
Fixed LPF/HPF (Generic)  & 84.12 & 86.41 & 54.02 \\
Fixed LPF/HPF (Learned)  & 84.45 & 86.50 & 54.31 \\
\rowcolor{bg}\textbf{FMA (Ours)}      & \textbf{85.24} & \textbf{86.73} & \textbf{55.66} \\
\bottomrule
\end{tabular}}
\label{tab:ablation_fma_adaptive_multidataset}
\end{table}

\begin{figure} [!ht]
    \centering 
    \includegraphics[width=0.5\linewidth]{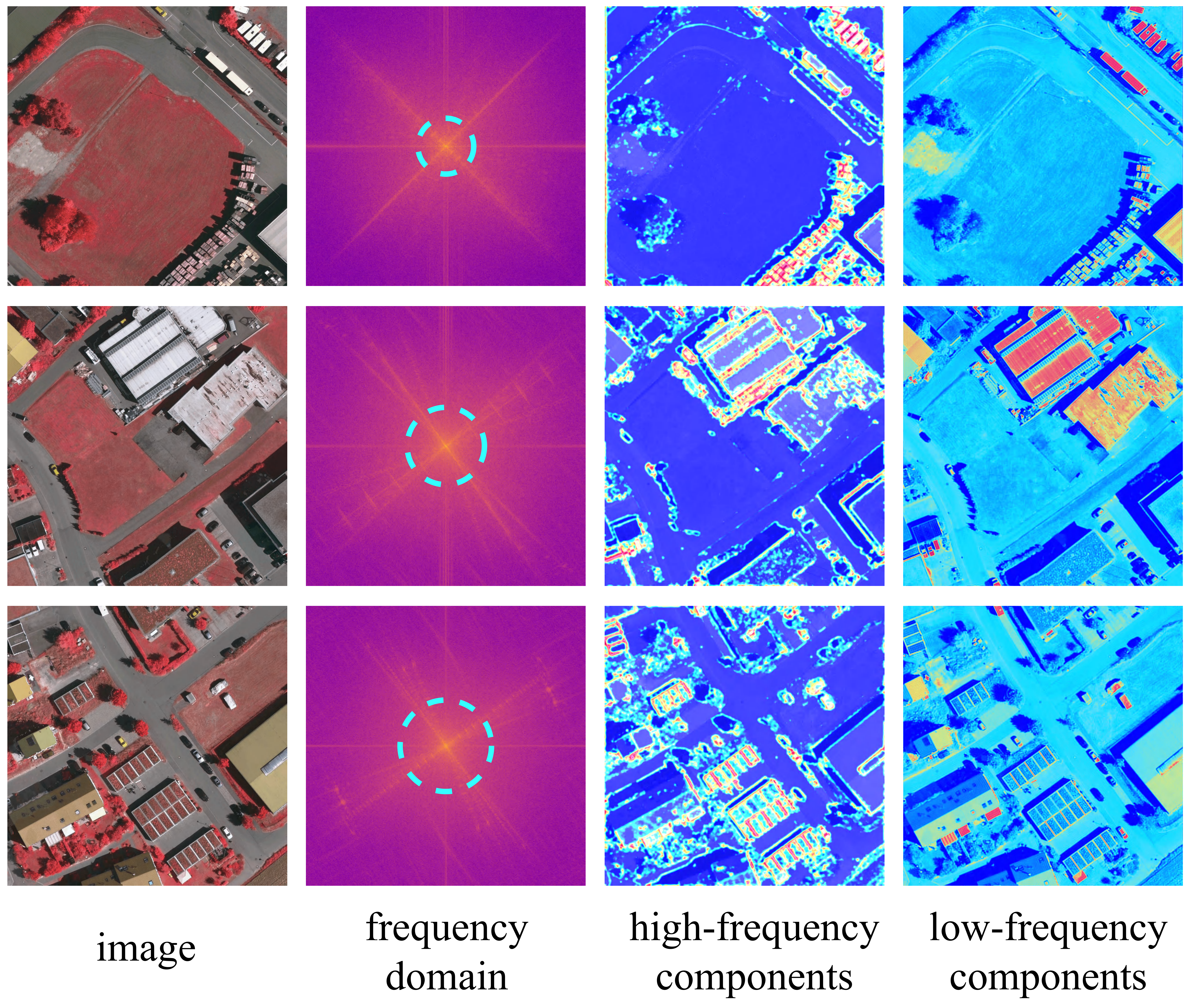}\\ 
    \caption{Visual validation of FMA's adaptive frequency partitioning on the ISPRS Vaihingen dataset.}
    \label{fig:fma_vis_vaih} 
\end{figure}

\begin{figure} [!ht]
    \centering 
    \includegraphics[width=0.5\linewidth]{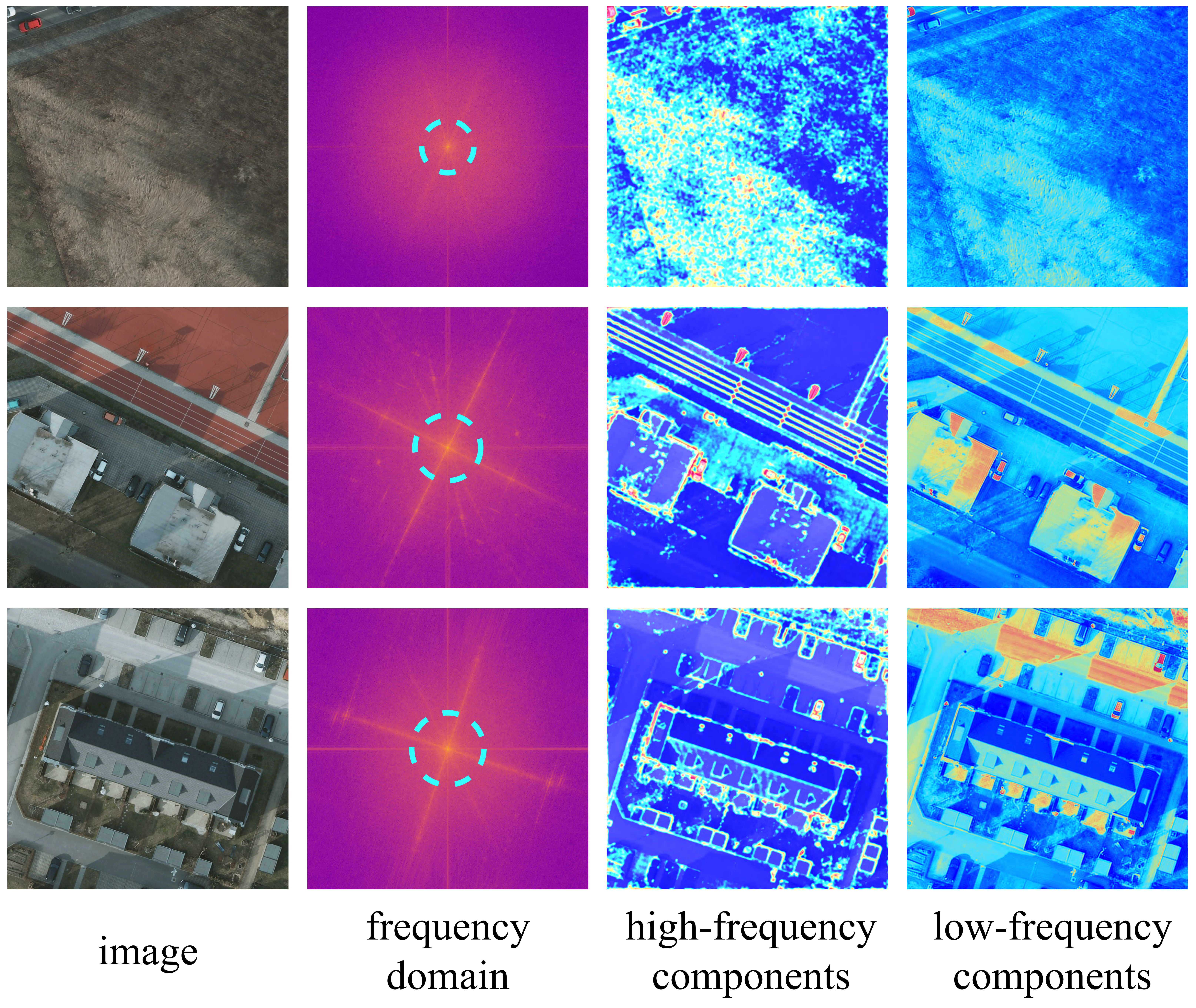}\\ 
    \caption{Visual validation of FMA's adaptive frequency partitioning on the ISPRS Potsdam dataset. }
    \label{fig:fma_vis_potsdam} 
\end{figure}

\textbf{Necessity of Adaptive Frequency Partitioning.} To prove that the effectiveness of adaptive frequency partitioning of FMA, we first quantitatively establish its superiority over conventional, non-adaptive frequency separation techniques in Table \ref{tab:ablation_fma_adaptive_multidataset}. Our adaptive FMA surpasses both the generic fixed filter and the learned static filter by a considerable margin across all three datasets. It confirms that a static, one-size-fits-all frequency cutoff is insufficient for the diverse characteristics of remote sensing scenes. We further visually validate how our FMA addresses the challenge of adapting SAM's features to diverse land cover types. As shown in Figs. \ref{fig:fma_vis_vaih} and  \ref{fig:fma_vis_potsdam}, the frequency partitioning is dictated by an adaptive radius $r$ (visualized as a dashed circle in the Fourier Domain), which is dynamically computed by accumulating spectral energy radially from the low-frequency origin until a predefined energy percentage $p$ is satisfied. As shown in Fig. \ref{fig:fma_vis_vaih}, when the image is full of intricate edges from buildings and cars, the FMA is compelled to integrate over a wider spectral area, systematically computing a larger radius $r$, a phenomenon consistently visible across these urban examples.


\subsection{Analysis of the Edge-Guided Refiner (EGRefiner)}

We conduct extensive experiments to verify the effectiveness of EGRefiner. The multi-scale feature extraction and different components are analyzed in detail. 

\textbf{Effectiveness of Multi-Scale Feature Extraction.} The design of EGRefiner is predicated on the hypothesis that a comprehensive edge representation for high-resolution remote sensing imagery requires fusing information from multiple receptive fields. We first validated this by comparing our multi-scale dilated convolution design against single-scale configurations. As shown in Table \ref{tab:egrefiner_multiscale}, the Multi-Scale DConv (Ours) approach achieves a superior mIoU of 85.24\% on the ISPRS Vaihingen dataset, outperforming all single-scale variants. This confirms that the synergistic aggregation of features is crucial for simultaneously capturing both fine-grained details via small kernels and larger structural information via larger kernels.

\begin{table}[htbp]
\centering
\caption{Ablation study on the multi-scale architecture of EGRefiner on three benchmark datasets (mIoU\%).}
\scalebox{0.9}{
\setlength{\tabcolsep}{3pt} 
\begin{tabular}{l|ccc}
\toprule
\textbf{Configuration} & \textbf{Vaihingen} & \textbf{Potsdam} & \textbf{LoveDA}\\
\midrule
Single-Scale DConv 3x3 & 84.72 & 86.55 & 54.71\\
Single-Scale DConv 5x5 & 84.85 & 86.60 & 55.06 \\
Single-Scale DConv 7x7 & 84.81 & 86.58 & 54.89\\
\rowcolor{bg}\textbf{Multi-Scale DConv (Ours)} & \textbf{85.24} & \textbf{86.73} & \textbf{55.66} \\
\bottomrule
\end{tabular}}
\label{tab:egrefiner_multiscale}
\end{table}

\begin{figure} [!ht]
    \centering 
    \includegraphics[width=3.5in]{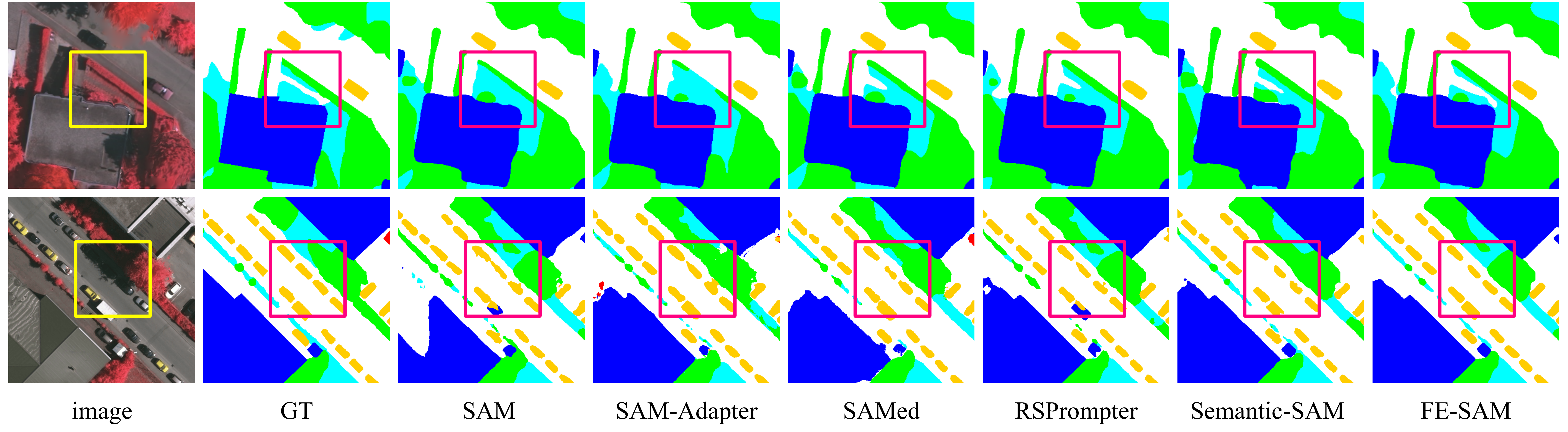} \\ 
    \caption{Qualitative comparison visualizing the efficacy of the Edge-Guided Refiner (EGRefiner) for boundary refinement on challenging scenes from the ISPRS Vaihingen dataset.}
    \label{fig:egrefiner_efficacy} 
\end{figure}

\textbf{Analysis of Critical Components in EGRefiner.} We conducted ablation study to validate the contribution of the EGRefiner's key components (Table \ref{tab:ablation_egrefiner_components_multidataset}). Removing the element-wise subtraction or replacing channel-wise concatenation with summation leads to a consistent and significant performance drop, confirming their importance in the EGRefiner. Furthermore, removing the fusion with the decoder's semantic features ($F_s$) causes a catastrophic drop in performance, with the mIoU falling far below even the baseline model without an EGRefiner. It is evident that low-level edge information is counterproductive when used in isolation. Without the high-level semantic context provided by $F_s$, the model is unable to distinguish between meaningful object boundaries and irrelevant high-frequency texture noise. 

\begin{table}[hbt!]
\centering
\caption{Ablation study on the architectural components of the EGRefiner, evaluated across all three benchmark datasets.}
\scalebox{0.9}{
\begin{tabular}{l|ccc}
\toprule
\textbf{EGRefiner Configuration} & \textbf{Vaihingen} & \textbf{Potsdam} & \textbf{LoveDA} \\
\midrule
w/o Element-wise Subtraction       & 84.65 & 86.48 & 55.03 \\
w/o Channel-wise Concat & 84.81 & 86.55 & 55.21 \\
w/o $F_s$ Feature Fusion & 79.32 & 81.05 & 49.58 \\
\rowcolor{bg}\textbf{EGRefiner (Full, Ours)}    & \textbf{85.24} & \textbf{86.73} & \textbf{55.66} \\
\bottomrule
\end{tabular}}
\label{tab:ablation_egrefiner_components_multidataset}
\end{table}

\textbf{Qualitative Analysis of Boundary Refinement.} We present visual comparisons in Fig. \ref{fig:egrefiner_efficacy} on challenging scenes from the ISPRS Vaihingen dataset. The results offer compelling visual evidence of the EGRefiner's critical contribution. Existing methods produce segmentation results with notably blurred edges. For instance, in the top row, the sharp corners of the building are rendered as smooth curves by the other methods. Similarly, in the bottom row, the other methods do not perform well in car segmentation. In contrast, our FE-SAM, empowered by the EGRefiner, delivers superior segmentation results. The visual comparisons validate that the EGRefiner has superior segmentation performance in fine-grained remote sensing applications.

\section{Conclusions}

In this paper, we proposed FE-SAM, a frequency- and edge-guided framework for adapting the Segment Anything Model to remote sensing image semantic segmentation. To alleviate the domain gap between natural and remote sensing images, the Frequency-Modulated Adapter performs input-adaptive frequency separation and prototype-based recalibration, enabling SAM to selectively enhance informative high- and low-frequency components with only a small number of trainable parameters. Meanwhile, the Edge-Guided Refiner introduces multi-scale structural cues from the original image to recover spatial details and improve object boundary delineation. Extensive experiments on the ISPRS Vaihingen, ISPRS Potsdam, and LoveDA datasets demonstrate that FE-SAM consistently outperforms state-of-the-art methods.

\bibliographystyle{IEEEtran}
\bibliography{ref}

\end{document}